\PassOptionsToPackage{table}{xcolor}
\documentclass[10pt,twocolumn,letterpaper]{article}

\usepackage{cvpr}

\usepackage{multirow}
\usepackage{makecell}
\usepackage{tcolorbox}
\tcbuselibrary{breakable}
\usepackage{cuted}
\usepackage{dblfloatfix}
\usepackage{float}
\usepackage{currfile}

\definecolor{lightblue}{RGB}{173,216,230}
\newcounter{prompt}
\renewcommand{\theprompt}{\arabic{prompt}}
\newcommand{\prompt}[3]{%
  \refstepcounter{prompt}%
  \begin{tcolorbox}[
    colback=lightblue!35,
    colframe=white!45!black,
    title={Prompt.~\theprompt:~#1},
    breakable,
  ]
  #2
  \label{#3}
  \end{tcolorbox}%
}

\definecolor{cvprblue}{rgb}{0.21,0.49,0.74}
\usepackage[pagebackref,breaklinks,colorlinks,allcolors=cvprblue]{hyperref}

\def\paperID{*****}
\def\confName{CVPR}
\def\confYear{2026}

\ifdefined\supplementarypaper
  \title{DualEraser: Joint Video Object and Effect Removal\\Supplementary Material}
\else
  \title{DualEraser: Joint Video Object and Effect Removal via Balanced Text-Mask Guidance and Decoupled Locator-Preserver}
\fi

\newcommand{\paperauthors}{%
  Yuqing Chen\textsuperscript{1,3} \quad
  Lin Liu\textsuperscript{2,*} \quad
  Haisu Wu\textsuperscript{4} \quad
  Xiaopeng Zhang\textsuperscript{2} \\
  Yaowei Wang\textsuperscript{5,3} \quad
  Yujiu Yang\textsuperscript{1,*} \quad
  Qi Tian\textsuperscript{2} \\
  \textsuperscript{1}Tsinghua University \quad
  \textsuperscript{2}Huawei \quad
  \textsuperscript{3}Pengcheng National Laboratory \\
  \textsuperscript{4}Southeast University \quad
  \textsuperscript{5}Harbin Institute of Technology%
}
\author{\paperauthors}

\begin{document}
\maketitle
\begingroup
\renewcommand{\thefootnote}{*}
\footnotetext{Corresponding authors.}
\endgroup

\begingroup
\preCutedStrip{\vskip-22.5pt}
\begin{strip}
\centering
\includegraphics[width=0.98\linewidth]{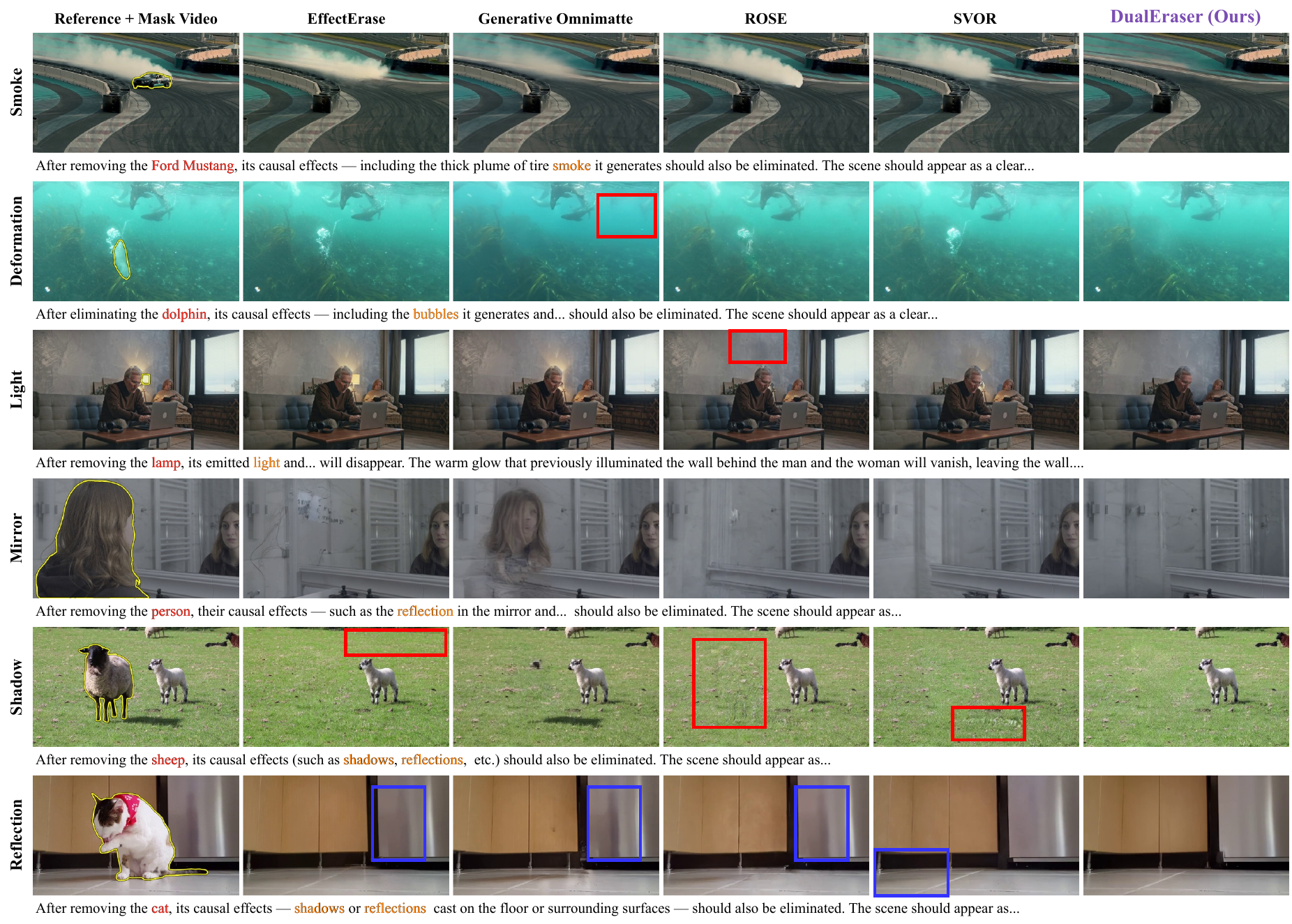}
\captionsetup{hypcap=false}
\captionof{figure}{Given the reference video, mask (yellow outline), and
Bipartite Text, DualEraser eliminates target objects (red text) and their
effects (yellow text). It outperforms existing methods by robustly removing
diverse complex effects.}
\captionsetup{hypcap=true}
\label{fig:main_exp}
\end{strip}
\endgroup

\begin{abstract}
Video object removal frequently struggles to eliminate target objects and their associated complex physical effects (e.g., smoke and light) in real-world scenes. We attribute this challenge to a fundamental \textbf{semantic--pixel conflict}, which manifests at two aspects: condition-level modality dissonance and optimization-level objective entanglement. In terms of conditioning, modality dissonance emerges from single-modality information incompleteness and  cross-modal dominance imbalance. During optimization, two conflicting objectives---high-level semantic erasure and  pixel-level background preservation---are inextricably entangled within a single model.
To address these conflicts, we propose DualEraser, a novel framework for joint video object and effect removal. First, a Bipartite Text prompt and a Multi-Conditional Capability Elicitation (MCCE) mechanism explicitly inject effect semantics and further leverage multimodal priors to address the limitations of individual modalities. Second, a Learnable Deep CFG Fusion (LD-CFG) module adaptively balances the relative dominance between the text and mask conditions. Finally, we introduce a decoupled expert architecture comprising a Locator for semantic erasure and a Preserver for background pixel alignment to break the objective entanglement. Extensive experiments demonstrate that DualEraser achieves state-of-the-art quantitative performance on standard benchmarks (e.g., gains of 2.16 dB and 1.44 dB on ROSE and VOR-Eval, respectively), while enabling  robust removal of complex effects in open-world videos. \href{https://cyqii.github.io/DualEraser.github.io/}{Project page}.
\end{abstract}

\section{Introduction}
Object removal, a technique increasingly indispensable in digital content creation and professional cinematography, aims to erase unwanted visual content while preserving scene integrity~\cite{he2025openve,mai2025easyv2v,cong2025viva,jiang2025vace,bian2025videopainter,wu2025qwen,yu2025anyedit}. In practice, however, conventional object-removal methods predominantly erase target objects while neglecting their associated physical effects~\cite{ekin2024clipaway,sun2025attentive,jiang2025smarteraser,li2025diffueraser,liu2025eraserdit,zi2025minimax,zhou2023propainter}, frequently leaving behind physically and visually inconsistent artifacts, such as residual shadows. To address this limitation, recent research has shifted toward the joint removal of objects and their physical effects~\cite{lee2025generative,samuel2025omnimattezero,miao2025rose,fu2026effecterase}. Despite this progress, achieving high-fidelity erasure of complex, weakly correlated effects, such as smoke (\cref{fig:main_exp}), remains a challenge. We attribute this challenge to a fundamental \textbf{semantic--pixel conflict} that operates at two levels: condition-level modality dissonance and optimization-level objective entanglement.

\begin{figure}[t]
    \centering
    \includegraphics[width=1.0\linewidth]{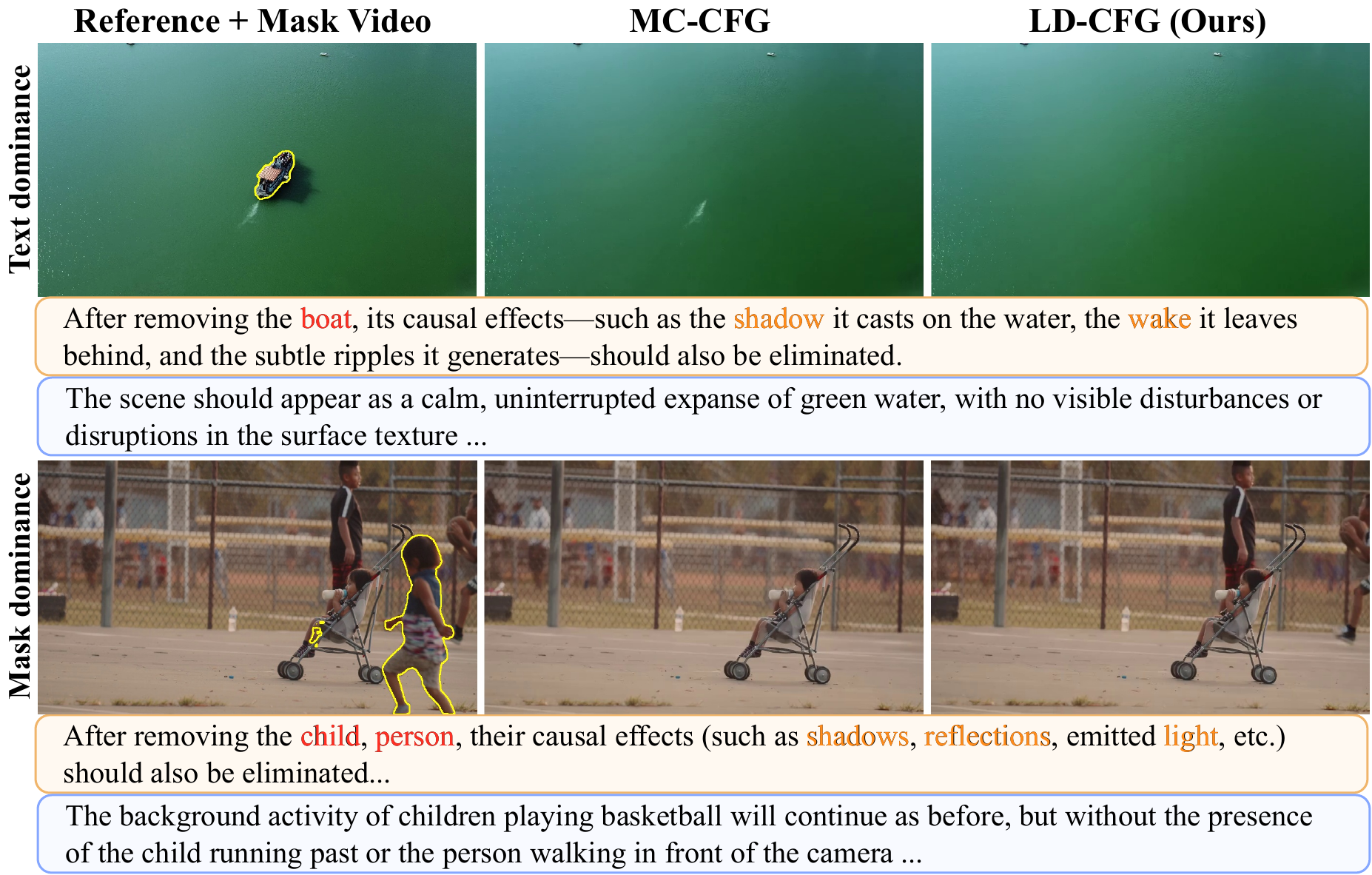}
    \caption{Text--mask dominance varies across scenarios. The bottom panel presents Bipartite Text, including removal targets (yellow), target objects and effects (colored terms), and the desired post-removal scene (blue). Row $1$ requires text-dominant guidance to remove ripples, whereas row $2$ requires mask-dominant guidance to preserve non-target individuals. LD-CFG adaptively balances the two modalities across various scenarios.}
    \label{fig:deep_cfg_ablation}
\end{figure}

At the condition level, modality dissonance first manifests as \emph{the information incompleteness of a single modality}. For example, pixel-level spatial masks lack adequate semantic depth about complex physical effects. Existing methods rely primarily on binary spatial masks, forcing models to implicitly infer object--effect correlations~\cite{miao2025rose,liu2025eraserdit,samuel2025omnimattezero,hu2026ideal,fu2026effecterase,lou2026learning,lee2025generative}. However, mask-only conditioning is unreliable for complex physical effects such as dispersed smoke and emitted light due to their weak spatiotemporal correlations with target objects.
In this context, explicit textual guidance can serve as a complementary modality to supply the semantics missing by leveraging the rich text-to-video priors of modern Multimodal Diffusion Transformer (MMDiT) models~\cite{wan2025wan,hacohen2024ltx,hacohen2026ltx,kong2024hunyuanvideo,yang2024cogvideox}.
Despite this potential, existing approaches exploit semantic guidance only in limited ways: ROSE~\cite{miao2025rose}, EraserDiT~\cite{liu2025eraserdit}, OmnimatteZero~\cite{samuel2025omnimattezero}, and SVOR~\cite{hu2026ideal} rely on static text prompts; EffectErase~\cite{fu2026effecterase} replaces textual object tokens with visual embeddings;  Learning Stochastic Bridges~\cite{lou2026learning} and Generative Omnimatte~\cite{lee2025generative} emphasize object-centric and background-centric descriptions, respectively. None of these approaches, however, specifies the complex physical effects to be removed.  To inject these critical missing semantics, we explicitly formulate the target effects via Bipartite Text conditions  (as illustrated in~\cref{fig:deep_cfg_ablation}). Coupled with our Multi-Conditional Capability Elicitation (MCCE), our approach fully exploits MMDiT priors to enhance the identification and elimination of complex effects.

While explicit textual guidance resolves the semantic deficit, combining it with spatial masks introduces another condition-level modality dissonance: \emph{cross-modal dominance imbalance}. As illustrated in \cref{fig:deep_cfg_ablation}, the relative importance of these two modalities shifts across scenarios due to their complementary nature—semantic prompts require spatial  masks for pixel-level precision, whereas spatial masks rely on the explicit Bipartite Text to supply semantic depth.
Existing work on multi-condition generation has focused primarily on coordinating visual control signals~\cite{wang2025dualreal,wang2025cinemaster}, leaving the balance between semantic text and spatial masks underexplored. In this context, Classifier-Free Guidance (CFG)~\cite{ho2022classifier} provides a natural mechanism for modulating the relative influence of text and other conditions through guidance-scale adjustment~\cite{ho2022classifier}. However, both  CFG~\cite{ho2022classifier} and its recent variants~\cite{papalampidi2025dynamic,gao2026c,kynkaanniemi2024applying} remain dependent on fixed scales, predefined schedules, or heuristically configured guidance rules, limiting their ability to achieve continuous, scene-adaptive balance.
To resolve this dissonance, we introduce a Learnable Deep CFG  (LD-CFG) fusion module. LD-CFG achieves scene-adaptive modality balance and mitigates the training-inference discrepancy inherent in traditional CFG strategies, ultimately yielding consistent performance gains.

\begin{figure}[t]

    \centering
    \includegraphics[width=1.0\linewidth]{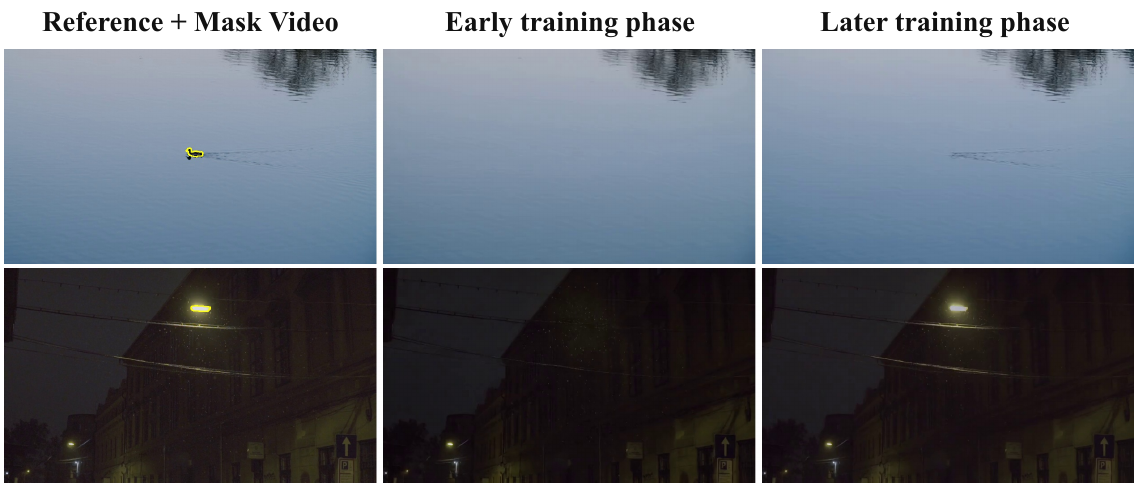}
    \caption{Qualitative comparisons of an expert model obtained at different phases of Stage I training on the VOR-Wild.}
    \label{fig:intro}
    \vspace{-0.5em}
\end{figure}

Beyond the condition-level dissonance, a more insidious \emph{optimization-level objective entanglement} plagues the architectural training phase. As shown in \cref{fig:intro} and \cref{tab:benchmark_training_phase}, an inherent optimization trade-off exists between high-level semantic erasure and  pixel-level background preservation. In the early training phase, models learn the global semantics to successfully remove target objects and their physical effects. However, as training progresses, they tend to overfit to local pixels to reconstruct the unedited regions. Since these two objectives are inherently entangled within a single network's weights, it is difficult for a single expert model to excel at both simultaneously.  Many existing methods~\cite{miao2025rose,fu2026effecterase,lou2026learning,lee2025generative} train a single expert. While they achieve high benchmark scores for pixel-level alignment, they sacrifice robust object elimination in open-world videos.
To break this entanglement, we introduce a decoupled expert architecture comprising a Locator and a Preserver.
The Locator is trained on multi-source data for fewer optimization steps to ensure consistent semantic erasure across scenes, enabling it to accurately locate and eliminate complex physical effects. In contrast, the Preserver is trained on more ``pixel-perfect'' datasets for an extended duration to maintain unedited-region  fidelity. Together, the two experts strike a balance between thorough elimination and precise background preservation.

By synergizing these targeted solutions, we present \textbf{DualEraser}, a comprehensive framework that breaks the semantic-pixel conflict for versatile and robust video object and effect removal. Overall, the main contributions of this work are summarized as follows:
\begin{itemize}
\item \textbf{Bipartite Text Guidance and Multi-Conditional Capability Elicitation.} We explicitly supply missing effect semantics via Bipartite Text guidance and exploit MMDiT priors through MCCE to improve the identification and removal of complex effects.
\item \textbf{Learnable Deep CFG  Fusion Module.} We introduce LD-CFG, a dynamic fusion module that adaptively balances text and mask guidance to resolve cross-modal dominance imbalance across scenes.
\item \textbf{Decoupled Expert Architecture. } We propose a Locator--Preserver architecture that decouples semantic erasure and pixel-level background preservation, thereby resolving optimization-level objective entanglement.
\item Extensive experiments demonstrate that DualEraser achieves state-of-the-art quantitative performance across standard benchmarks while delivering  robust complex effect elimination in open-world videos.
\end{itemize}

\begin{table}[t]
\centering

\scriptsize
\setlength{\tabcolsep}{1pt}
\renewcommand{\arraystretch}{1.2}
\resizebox{1.0\columnwidth}{!}{
\begin{tabular}{l|c|c|c}
\toprule
\multirow[c]{2}{*}{Training phase}
& ROSE Bench
& VOR-Eval
& VOR-Wild \\
& PSNR $\uparrow$
& PSNR $\uparrow$
&EPR \\
\hline\hline
Early phase
& 30.39
& 23.29
& \textbf{0.4987} \\

Late phase
& \textbf{32.86}
& \textbf{23.53}
& 0.4346 \\
\bottomrule
\end{tabular}
}
\caption{Performance evaluation of a single expert model at early and late phases of Stage I training on the ROSE Benchmark, VOR-Eval and VOR-Wild.}
\label{tab:benchmark_training_phase}
\end{table}

\section{Related Work}
\subsection{Video Object Removal}
Most existing object-removal methods focus on erasing the mask-specified target object, without  modeling the physical effects induced by it. Image-domain methods have achieved strong performance in object removal~\cite{ekin2024clipaway, sun2025attentive, jiang2025smarteraser}, while video-domain methods further address temporal consistency~\cite{zhou2023propainter,li2025diffueraser,liu2025eraserdit,zi2025minimax}.  However, removing the target object while leaving its associated physical effects, such as shadows and reflections, often produces physically inconsistent scenes. Recent studies therefore explore joint object-and-effect removal. Image-based methods address static effects~\cite{wei2025omnieraser, zhu2025georemover, zhao2025objectclear}, whereas video-based methods must additionally preserve temporal coherence. Specifically, Generative Omnimatte~\cite{lee2025generative} fine-tunes diffusion models for removal, while OmnimatteZero~\cite{samuel2025omnimattezero} achieves training-free removal via spatio-temporal attention guidance. ROSE~\cite{miao2025rose} leverages a difference mask predictor for precise inpainting. EffectErase~\cite{fu2026effecterase} proposes an auxiliary target-insertion task to enhance removal performance. Furthermore, SVOR~\cite{hu2026ideal} addresses removal under imperfect conditions, and Learning Stochastic Bridges~\cite{lou2026learning} treats it as a video-to-video translation problem. Overall, despite their respective advancements, these approaches often fall short in removing complex physical effects in in-the-wild videos.
\begin{figure*}[t]
    \centering
    \includegraphics[width=1.0\linewidth]{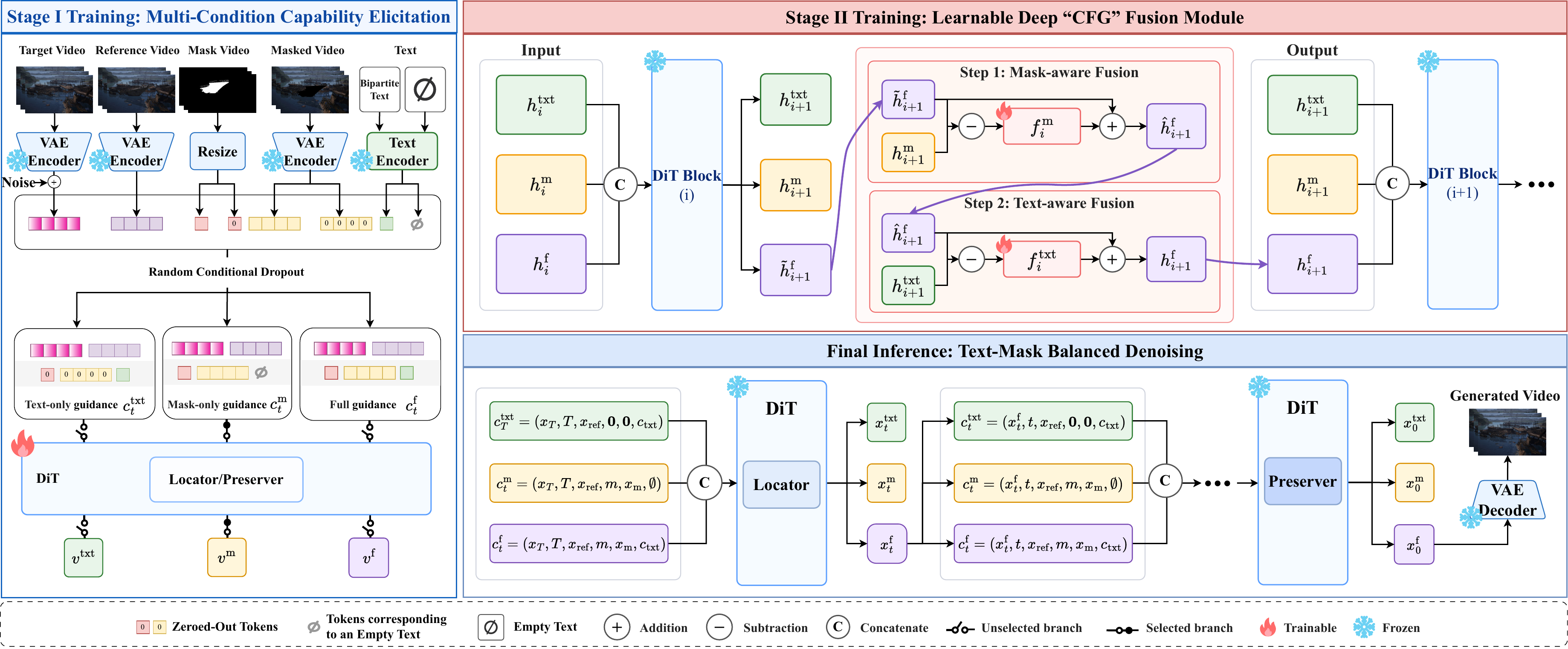}
    \caption{The two-stage training framework of DualEraser. During Stage I training, the MCCE  randomly selects one of three conditioning configurations for optimization to enhance model alignment with different conditions. During Stage II training,  LD-CFG  integrates features from three deterministic conditioning paths to  balance text and mask modalities.  Finally, during the inference phase, only the output corresponding to the full guidance is utilized for the subsequent denoising process.}
    \label{fig:framework}
\end{figure*}

\subsection{Balancing Textual and Control Conditions}
Most recent video removal models are built on MMDiT backbones~\cite{wan2025wan,hacohen2024ltx,hacohen2026ltx,kong2024hunyuanvideo,yang2024cogvideox}, which provide pretrained text-to-video priors and support multiple conditioning signals, such as text prompts and spatial masks ~\cite{wan2025wan, yang2024cogvideox, hacohen2024ltx,kong2024hunyuanvideo}. In video object-and-effect removal, these conditions play complementary roles: text guidance provides semantic cues for object-induced effects, whereas mask-based control offers spatial precision. Since the relative dominance of text and mask guidance varies across scenes, scene-adaptive balancing is essential. Existing studies primarily focus on coordinating multiple visual control signals~\cite{wang2025cinemaster,wang2025dualreal}, whereas the balance between text and mask guidance remains underexplored.
Under these circumstances, CFG~\cite{ho2022classifier} provides a foundational mechanism to weight text against other conditions via guidance scale adjustments. To refine this mechanism, recent efforts have introduced various strategies: Dynamic CFG~\cite{papalampidi2025dynamic} employs an online latent-space evaluator for step-wise adjustment, C$^2$FG~\cite{gao2026c} introduces an exponential decay function, and Limited Interval Guidance~\cite{kynkaanniemi2024applying} restricts CFG to specific noise-level intervals. Despite these advancements, these methods still  rely on manual, heuristic tuning to define their operational hyperparameters. Consequently, existing methods struggle to adaptively balance text and mask guidance for joint object-and-effect removal in real-world videos.

\subsection{Mixture-of-Experts for Visual Generation and Editing}
Building on the success of MoE architectures in large language models~\cite{fedus2022switch,dai2024deepseekmoe,liu2024deepseek}, recent studies have extended MoE designs to visual generation and editing. Within diffusion backbones, existing MoE applications can be broadly categorized into two design patterns: denoising-stage specialization and token-level expert routing. Under the former design pattern, Wan2.2~\cite{wan2025wan} adopts a dual-expert architecture that partitions the denoising trajectory by noise level: the high-noise expert synthesizes global layouts and contours, whereas the low-noise expert refines fine-grained textures and details.
Under the latter design pattern, methods employ token-level expert routing to improve the scalability and efficiency of DiT architectures. For instance, DiT-MoE~\cite{fei2024scaling} uses sparse expert routing to process spatial tokens efficiently. DiffMoE~\cite{shi2025diffmoe} further introduces a batch-level global token pool and a dynamic capacity predictor to improve token routing in diffusion transformers.
Beyond generative-model scaling, MoE architectures have also been applied to downstream editing tasks, enabling task-specific specialization and adaptive routing of conditional signals. OpenVE-Edit~\cite{he2025openve} uses a task-aware MoE connector to accommodate diverse editing tasks efficiently. Similarly, AnyEdit~\cite{yu2025anyedit} integrates MoE modules into U-Net cross-attention layers to allocate visual conditions to the most relevant experts. Although these methods apply MoE to diverse visual tasks, none has explored its use in decoupling semantic erasure from background preservation for video object-and-effect removal.

\section{Method}
DualEraser is trained using a two-stage procedure, as shown in \cref{fig:framework}. During Stage I, we train MCCE to strengthen the model's alignment with text and mask conditions, thereby improving its overall erasure capability. During Stage II, we train LD-CFG using three deterministic conditioning paths to adaptively balance the relative dominance of text and mask guidance across scenes. The Locator and Preserver independently follow this two-stage procedure with expert-specific training configurations and noise levels. Finally, during inference, only the full-guidance output is propagated to the next denoising step.

\subsection{Multi-Conditional Capability Elicitation}
For video object and effect removal, we introduce an MCCE mechanism to more effectively activate and exploit modality-specific generative priors.

As \cref{fig:framework} illustrates, given a reference video $V_{\text{ref}}$ and a mask $M$, the masked video is $V_{\text{m}} = V_{\text{ref}} \odot (\mathbf{1} - M)$. Let $x_{\text{ref}}$, $m$, and $x_{\text{m}}$ denote their respective latents, and let $x_0$ denote the latent of the ground-truth target video $V_{\text{gt}}$. Following the flow-matching formulation of the Wan backbone~\cite{wan2025wan},
the interpolated latent and its target velocity field are defined as
\begin{equation}
\begin{aligned}
    x_t &= (1-t)x_0+t\epsilon, \quad t\in[0,1],\\
    v_t^\star &= \frac{\partial x_t}{\partial t}
    =\epsilon-x_0.
\end{aligned}
\label{eq:flow_matching_velocity}
\end{equation}
where $\epsilon\sim\mathcal{N}(0,I)$ denotes standard Gaussian
noise, $t$ is the flow-matching timestep, and $v_t^\star$ is the
target flow-matching velocity (vector field).

To specify what should be removed and the desired post-removal scene, we condition the model on a Bipartite Text prompt $\mathcal{P}$, encoded as $c_{\text{txt}}$. Unlike conventional text prompts, $\mathcal{P}$ explicitly specifies the target object, its associated physical effects, and the desired post-removal scene (\cref{fig:deep_cfg_ablation}). This formulation supplies the missing semantic cues needed to identify and remove complex physical effects.

During Stage I, as illustrated in \cref{fig:framework}, we apply conditional dropout to encourage modality-specific capability elicitation.  Building on the CFG mechanism used in MMDiT, which randomly drops the text condition during training, we extend the conditional-dropout strategy by randomly zeroing the mask latent $m$ and the masked-video latent $x_{\text{m}}$. Consequently, the model is optimized under three conditioning configurations: (1) text-only guidance, where $c^\text{txt}_t=(x_t,t, x_\text{ref},\mathbf{0},\mathbf{0},c_{\text{txt}})$; (2) mask-only guidance, where $c^\text{m}_t=(x_t, t, x_\text{ref},m,x_\text{m},\emptyset)$, with $\emptyset$ denoting the empty text latent; and (3) joint (full) guidance, where $c^\text{f}_t=(x_t, t, x_\text{ref},m,x_\text{m},c_{\text{txt}})$. The text-only configuration compels the model to reconstruct targets using solely textual cues, strengthening text-to-visual alignment. Notably, the unconditional case is excluded, as its inherent stochasticity conflicts with the deterministic control required for removal tasks~\cite{wolf2026continuous}.

\textbf{MC-CFG Baseline Inference.}
To enable a fair comparison with our LD-CFG module trained during Stage II, we extend conventional CFG to a multi-conditional inference baseline (MC-CFG) that accommodates both text and mask conditions. The final flow-matching velocity $v_t$ is obtained through three sequential operations. First, the mask guidance scale $\omega_{\text{m}}$ controls extrapolation along the direction $v^{\text{f}}_t-v^{\text{m}}_t$, moving from the mask-only prediction toward the full-condition prediction. The resulting prediction is then rescaled to prevent signal over-amplification. Finally, the text guidance scale $\omega_{\text{txt}}$ controls fusion along the direction $\hat{v}_t-v^{\text{txt}}_t$, moving from the text-only prediction toward the rescaled intermediate prediction:
\begin{equation}
\begin{aligned}
    \tilde{v}_{t} &= v^{\text{m}}_{t} + \omega_{\text{m}} \left( v^{\text{f}}_{t} - v^{\text{m}}_{t} \right), \\
    \hat{v}_{t} &= \operatorname{clip} \left( \frac{\lVert v^{\text{f}}_{t} \rVert_{2}}{\lVert \tilde{v}_{t} \rVert_{2} + \delta}, 0, 1 \right) \tilde{v}_{t}, \\
    v_{t} &= v^{\text{txt}}_{t} + \omega_{\text{txt}} \left( \hat{v}_{t} - v^{\text{txt}}_{t} \right),
\end{aligned}
\end{equation}
where $v^{\text{m}}_{t}$, $v^{\text{txt}}_{t}$, and $v^{\text{f}}_{t}$ denote the flow-matching velocity predictions under mask-only, text-only, and joint full conditioning, respectively, and $\delta = 10^{-8}$ ensures numerical stability.

\subsection{Learnable Deep ``CFG'' Fusion Module}
While MC-CFG balances text and mask guidance using manually specified inference-time scales, these static scales cannot adapt to scene-dependent shifts in modality dominance. Accordingly, we introduce the LD-CFG module to adaptively balance text and mask guidance across scenes. As \cref{fig:framework} illustrates, the second training stage replaces random conditional dropout with a concurrent, deterministic three-path input strategy: (1) text-only, (2) mask-only, and (3) joint guidance. Specifically, the $i$-th DiT block processes inputs along these three conditioning paths in parallel:
\begin{equation}
  [h^\text{txt}_{i+1}; h^\text{m}_{i+1}; \tilde{h}^{\text{f}}_{i+1}] = \text{Block}_i([h^\text{txt}_{i}; h^\text{m}_{i}; {h}^{\text{f}}_{i}]),
\end{equation}
where $h^\text{txt}_{i}$, $h^\text{m}_{i}$, and ${h}^{\text{f}}_{i}$ denote the inputs from the text-only, mask-only, and joint branches, respectively. The operator $\text{Block}_i(\cdot)$ yields updated representations for the isolated conditions ($h^\text{txt}_{i+1}$ and $h^\text{m}_{i+1}$) alongside an intermediate pre-fusion representation $\tilde{h}^{\text{f}}_{i+1}$ for the joint branch.

To adaptively integrate modality-specific cues, we apply  learnable linear projections after each DiT block to fuse the features produced along the three conditioning paths.
\begin{equation}
  \begin{aligned}
  \hat{h}^{\text{f}}_{i+1} &= \tilde{h}^{\text{f}}_{i+1} + f^\text{m}_{i}( \tilde{h}^{\text{f}}_{i+1} - h^\text{m}_{i+1}),\\
  {h}^{\text{f}}_{i+1} &= \hat{h}^{\text{f}}_{i+1} + f^\text{txt}_{i}(\hat{h}^{\text{f}}_{i+1} - h^\text{txt}_{i}),
  \end{aligned}
\end{equation}
where $f^\text{m}_{i}$ and $f^\text{txt}_{i}$ denote the learnable linear projections  corresponding to the mask-aware and text-aware fusion, respectively. In this formulation, $\hat{h}^{\text{f}}_{i+1}$ represents the intermediate state of the joint branch after assimilating the mask-guided residual, whereas ${h}^{\text{f}}_{i+1}$ denotes the final fused feature. Finally, the current block reconstructs its output as $[h^\text{txt}_{i+1}; h^\text{m}_{i+1}; {h}^{\text{f}}_{i+1}]$ and passes it to the subsequent DiT block.

\textbf{Proposed LD-CFG Inference.} In contrast to the comparative baseline discussed above, our proposed method adopts an adaptive inference strategy. During inference, as illustrated in \cref{fig:framework}, our method propagates only the output of the full-guidance  path to the next denoising step:
\begin{equation}
    d x_t = v^{\mathrm{f}}_t\,dt,
\end{equation}
where the  full-conditioning vector field is defined by
\begin{equation}
    v^{\mathrm{f}}_t
    :=
    \Pi_{\mathrm{full}}
    \left[
    v_\theta
    \left(
    [c^{\text{txt}}_t,c^{\text{m}}_t,c^{\text{f}}_t]
    \right)
    \right].
\end{equation}
Here, $[c^{\text{txt}}_t,c^{\text{m}}_t,c^{\text{f}}_t]$ denotes batch-wise concatenation, $v_\theta(\cdot)$ is the denoising model, and $\Pi_{\mathrm{full}}[\cdot]$ extracts the third batch chunk representing the full branch. Ultimately, this formulation achieves an adaptive balance across disparate conditions while  exploiting pre-trained generative priors. After  $N$ denoising steps, the clean video latent $x^{\text{f}}_0$ is generated and processed by the VAE
decoder to reconstruct the output video.

\begin{table*}[t]
\centering

\tiny
\setlength{\tabcolsep}{3pt}
\renewcommand{\arraystretch}{0.95}
\resizebox{0.8\linewidth}{!}{
\begin{tabular}{l|ccc|ccc|cc}
\toprule
\multirow[c]{2}{*}{Method}
& \multicolumn{3}{c|}{ROSE Bench}
& \multicolumn{3}{c|}{VOR-Eval}
& \multicolumn{2}{c}{VOR-Wild} \\
& PSNR$\uparrow$ & SSIM$\uparrow$ & LPIPS$\downarrow$
& PSNR$\uparrow$ & SSIM$\uparrow$ & LPIPS$\downarrow$
& EPR$\uparrow$ & UPR$\uparrow$ \\
\hline\hline
MiniMax
& 26.19 & 0.8978 & 0.0768
& 21.07 & 0.7619 & 0.1519
& -- & -- \\

DiffuEraser
& 26.60 & 0.8995 & 0.0930
& 21.25 & \underline{0.7729} & 0.1534
& -- & -- \\

ProPainter
& 27.04 & 0.9179 & 0.0662
& 21.00 & 0.7712 & 0.1496
& -- & -- \\
\hline

Gen. Omnimatte
& 27.03 & 0.8922 & 0.0866
& 21.73 & 0.7589 & 0.1537
& 0.4103 &	0.4872 \\

EffectErase
& 27.04 & 0.8987 & 0.0679
& 22.47 & 0.7467 & 0.1296
& 0.4256 & 0.4462\\

ROSE
& 31.09 & 0.9181 & 0.0527
& 22.07 & 0.7606 & 0.1402
& 0.3744 & 0.3744\\

SVOR
& 31.17 & 0.9163 & 0.0552
& 21.91 & 0.7565 & 0.1384
& 0.4769 & 0.5179 \\
\hline

\rowcolor{cyan!10}
DualEraser 1.3B
& \underline{32.21} & \textbf{0.9262} & \underline{0.0483}
& \underline{23.08} & \textbf{0.7738} & \underline{0.1271}
& \underline{0.5410} & \underline{0.5538}\\

\rowcolor{cyan!10}
DualEraser 5B
& \textbf{33.33} & \underline{0.9256} & \textbf{0.0461}
& \textbf{23.91} & 0.7708 & \textbf{0.1194}
& \textbf{0.5872} & \textbf{0.5744} \\
\bottomrule
\end{tabular}
}
\caption{Quantitative comparisons of various methods on the ROSE Benchmark, VOR-Eval, and VOR-Wild datasets.}
\label{tab:method_comparison}
\label{main-tab:method_comparison}
\end{table*}

\subsection{Locator and Preserver: Decoupling Semantic Erasure and Pixel Preservation}
As illustrated in \cref{fig:intro} and \cref{tab:benchmark_training_phase}, a single expert model faces a critical dilemma: while metrics on standard benchmarks improve, the model's erasure capabilities in open-world scenarios degrade. This degradation results from prolonged MSE-based optimization, which shifts the model from learning removal semantics to overfitting pixel-level reconstruction, thereby limiting its generalization to open-world videos. To decouple these two competing objectives, we define two experts operating at different noise levels: the Locator (high-noise expert) identifies and eliminates target objects and effects, while the Preserver (low-noise expert) ensures faithful background preservation.

Furthermore, it is observed that the two experts possess  distinct data requirements. The Locator benefits from diverse synthetic and real-world data, which strengthens its ability to recognize objects and associated effects in complex environments. Conversely, the Preserver requires pixel-level correspondence between input and target videos to learn faithful background preservation. Real-world pairs (e.g., VOR~\cite{fu2026effecterase}) often exhibit background misalignment due to imperfect data collection, causing the model to fit spurious artifacts. Synthetic data is therefore better suited to this requirement, as it provides cleaner pixel-level supervision.

Notably, our proposed Locator and Preserver design differs from the generative noise-level expert  of Wan2.2~\cite{wan2025wan} along three key dimensions: (1) training objectives, targeting semantic erasure and background preservation rather than layout formation and detail refinement; (2) training duration, with fewer steps for the Locator to retain high-level semantic erasure and longer optimization for the Preserver to improve pixel-level alignment; and (3) training data composition, using diverse synthetic and real-world data for the Locator and ``pixel-perfect'' data for the Preserver.

\section{Experiments}
\paragraph{Experimental Implementation}
DualEraser is built upon  Wan2.2 5B and Wan2.1 1.3B~\cite{wan2025wan}, with Wan2.2 5B serving as the default model unless otherwise specified.
For our training data, we utilize two primary datasets: VOR~\cite{fu2026effecterase} containing  60k video pairs, and ROSE~\cite{miao2025rose} containing approximately 17k video pairs.
In Stage I training, we set the text-condition dropout probability to 0.1 and the probability of jointly zeroing the mask and masked-video latents to 0.2. For the 5B model, the Locator and Preserver are trained for 12,500 and 20,000 steps, respectively; for the 1.3B model, they are trained for 15,000 and 20,000 steps. The Locator uses the balanced VOR--ROSE mixture, whereas the Preserver uses the rebalanced ROSE dataset (see the Appendix for sampling details). Stage I therefore comprises $32.5\mathrm{K}$ and $35\mathrm{K}$ expert-training steps for the 5B and 1.3B models, respectively. We set the noise-level boundary between the two experts to 0.875. In Stage II, for both model sizes, we independently optimize the Locator and Preserver on the balanced VOR--ROSE mixture for an additional 1,800 steps each, corresponding to $3.6\mathrm{K}$ expert-training steps in total. All training is conducted with a batch size of 2 across 16 A100 GPUs. At inference, we use 40 denoising steps. In addition, the Bipartite Text prompts for training and evaluation are generated using Qwen3-VL 8B~\cite{bai2025qwen3}.

\paragraph{Baselines}
The  isolated object removal baselines include ProPainter~\cite{zhou2023propainter}, MiniMax-Remover (MiniMax)~\cite{zi2025minimax}, and DiffuEraser~\cite{li2025diffueraser}. Furthermore, the joint object-and-effect removal baselines comprise ROSE~\cite{miao2025rose}, EffectErase~\cite{fu2026effecterase}, SVOR~\cite{hu2026ideal}, and Generative Omnimatte (Gen. Omnimatte)~\cite{lee2025generative}.

\paragraph{Benchmarks and Metrics}
For quantitative evaluation, the ROSE Benchmark~\cite{miao2025rose} and VOR-Eval~\cite{fu2026effecterase} are adopted, both of which provide ground-truth videos. The corresponding evaluation metrics include PSNR~\cite{hore2010image}, SSIM~\cite{wang2004image}, and LPIPS~\cite{zhang2018unreasonable}. To assess removal capabilities on in-the-wild videos, the VOR-Wild~\cite{fu2026effecterase} is utilized as the primary dataset. For this benchmark, the Erasure Preference Rate (EPR) is employed as the evaluation metric. In each experimental group, the generated videos are anonymized and randomly distributed to two human experts specialized in removal tasks. These experts are instructed to select multiple samples in which both the target objects and their  effects are successfully eliminated while preserving high video quality. The methods responsible for generating the selected samples are assigned a score of $1$, whereas the unselected methods receive a score of $0$. The final EPR for a given method is then calculated as its average score across all evaluated samples. Furthermore, the User Preference Rate (UPR) is determined by recruiting five volunteers to assess the various samples following an identical protocol.

\begin{table}[t]
\centering

\setlength{\tabcolsep}{6pt}
\begin{tabular}{c|cccc}
\toprule
\multirow[c]{2}{*}{\makecell{Locator \\ Training Step} }
& \multicolumn{4}{c}{EPR} \\
\cline{2-5}
& $p=0$ & $p=0.2$ & $p=0.5$ & $p=0.8$ \\
\midrule
1000* & 0.3564 & 0.3615 & 0.3667 & \textbf{0.4051} \\
2000* & {0.3692} & \textbf{0.4000} & {0.3692} & 0.3744 \\
3000* & 0.3436 & \textbf{0.3923} & {0.3692} & 0.3205 \\
4000* & 0.3385 & \textbf{0.3718} & 0.3462 & 0.3590 \\
\bottomrule
\end{tabular}
\caption{Performance on VOR-Wild across different Locator training steps and conditional dropout probabilities $p$ during Stage I training. All configurations share a fixed Preserver trained on ROSE. The asterisk (*) denotes training exclusively on ROSE.}
\label{tab:probabilities}
\end{table}

\subsection{Comparison with State-of-the-Art Methods}
\paragraph{Quantitative results}  As reported in \cref{tab:method_comparison}, both DualEraser variants achieve the best or second-best results on most metrics. In particular, the 5B variant improves PSNR by $2.16$ dB over SVOR on the ROSE Benchmark and by $1.44$ dB over EffectErase on VOR-Eval. It also attains the highest EPR and UPR on VOR-Wild.
Together, these results demonstrate that DualEraser not only achieves superior quantitative performance on standard benchmarks but also delivers more robust object-and-effect removal on in-the-wild videos.

\paragraph{Qualitative results} As shown in \cref{fig:main_exp}, DualEraser effectively removes diverse physical effects, including smoke, deformation, emitted light, mirrors, shadows, and reflections. In contrast, competing methods exhibit background discontinuities (red boxes) or incomplete erasure (blue boxes). Specifically, Gen. Omnimatte introduces an unnatural blue tint despite removing the bubbles, while SVOR leaves noticeable white artifacts.

\begin{table}[t]
\centering

\scriptsize
\setlength{\tabcolsep}{1pt}
\renewcommand{\arraystretch}{0.98}
\resizebox{0.9\linewidth}{!}{
\begin{tabular}{l|c|c|c}
\toprule
\multirow[c]{2}{*}{Model}
& ROSE Bench
& VOR-Eval
& VOR-Wild \\
& PSNR$\uparrow$
& PSNR$\uparrow$
& EPR$\uparrow$ \\
\hline\hline
Average MC-CFG
& 32.69
& 23.69
& -- \\

Standard MC-CFG
& 32.91
& 23.79
& 0.5154 \\

Simple Linear
& 32.16
& 23.41
& 0.5077 \\

\rowcolor{cyan!10}
LD-CFG (Ours)
& \textbf{33.33}
& \textbf{23.91}
& \textbf{0.5846} \\
\bottomrule
\end{tabular}
}
\caption{Ablation study of LD-CFG during Stage II Training. Average MC-CFG denotes average performance across 16 combinations of text ($\omega_{\text{txt}} \in \{1.0, 1.5, 2.0, 3.0\}$) and mask ($\omega_{\text{m}} \in \{1.0, 1.5, 2.0, 3.0\}$) guidance scales. Standard MC-CFG uses the default setting. Simple Linear appends a linear layer directly to each block output and matches the parameter count of LD-CFG.}
\label{tab:deep_cfg}
\label{main-tab:deep_cfg}
\end{table}

\subsection{Ablation Studies and Analysis}
To  validate the effectiveness of  DualEraser, comprehensive ablation studies are conducted to evaluate the individual contributions of  the MCCE mechanism, the LD-CFG module, and the Locator–Preserver architecture.

\paragraph{Multi-Conditional Capability Elicitation}
To accelerate convergence, we train the Locator exclusively on ROSE for this ablation. As shown in \cref{tab:probabilities}, $p=0$ disables the proposed mask-condition dropout and serves as the Wan2.2-based baseline~\cite{wan2025wan}, whereas $p>0$ jointly zeros the mask and masked-video latents with probability $p$. Across most training durations, nonzero $p$ values improve EPR over $p=0$. These results support the effectiveness of conditional dropout within MCCE for in-the-wild erasure.

\paragraph{Learnable Deep CFG Fusion Mechanism}
As shown in \cref{tab:deep_cfg}, LD-CFG (row $4$) outperforms both average and standard MC-CFG (rows $1$ and $2$) on all reported metrics, demonstrating the benefit of scene-adaptive guidance fusion over fixed guidance scales. More importantly, its advantage over Simple Linear (row $3$) further indicates that the gains arise from the proposed fusion design rather than additional parameters alone.

\paragraph{Locator and Preserver}
\Cref{tab:probabilities,tab:locator_rose,tab:Preserver_comparison} reveal contrasting training dynamics for the Locator and Preserver.
For the Locator, while EPR generally declines from 2,000 to 4,000 steps across tested dropout probabilities (\Cref{tab:probabilities}), longer training improves some reconstruction metrics (\Cref{tab:locator_rose}). This indicates a clear trade-off between pixel-level fitting and robust semantic erasure. In contrast, longer Preserver training improves $\text{PSNR}^{\text{ref}}_{\text{bg}}$ while maintaining relatively stable EPR (see the blue rows in \Cref{tab:Preserver_comparison}).

\Cref{tab:Preserver_comparison} further examines the effect of training data composition on the Preserver. Training exclusively on ROSE yields greater reference-background fidelity than training on the VOR--ROSE mixture. In particular, even after only 5,000 steps, training solely on the ROSE dataset yields a higher $\text{PSNR}^{\text{ref}}_{\text{bg}}$ than training the Preserver on a mixed VOR and ROSE dataset for 40,000 steps, suggesting that pixel-aligned data is more suitable for the Preserver. Together, these results support our decoupled design: the Locator prioritizes robust semantic erasure, whereas the Preserver specializes in faithful background reconstruction.

\begin{table}[t]
\centering

\begin{tabular}{c|ccc}
\toprule
\multirow{2}{*}{\makecell{Locator\\Training Step}}
& \multicolumn{3}{c}{ROSE Bench} \\
\cline{2-4}
& PSNR$\uparrow$ & SSIM$\uparrow$ & LPIPS$\downarrow$ \\
\hline\hline
1000*
& 31.87 & 0.9199 & 0.0517 \\

2000*
& 32.48 & 0.9219 & 0.0499 \\

3000*
& \textbf{32.51} & \textbf{0.9227} & 0.0501 \\

4000*
& 32.22 & 0.9193 & \textbf{0.0479} \\
\bottomrule
\end{tabular}
\caption{Ablation study of Locator during Stage I Training.  Across all configurations, an identical Preserver is utilized, which was trained on  ROSE dataset. The asterisk (*) denotes training exclusively on the ROSE training dataset.}
\label{tab:locator_rose}
\end{table}

\begin{table}[t]
\centering
\scriptsize
\setlength{\tabcolsep}{1pt}
\renewcommand{\arraystretch}{0.9}
\resizebox{0.45\textwidth}{!}{
\begin{tabular}{l|c|cc|c}
\toprule
\multirow[c]{2}{*}{\makecell{Preserver\\Training Step}}
& \multicolumn{1}{c|}{ROSE Bench}
& \multicolumn{2}{c|}{VOR-Eval}
& \multicolumn{1}{c}{VOR-Wild} \\
& PSNR$\uparrow$
& PSNR$\uparrow$
& $\text{PSNR}^{\text{ref}}_{\text{bg}}\uparrow$
& EPR \\
\hline\hline
20000
& 31.98
& 23.51
& 51.57
& 0.5077 \\

30000
& 31.85
& 23.69
& 52.21
& 0.4949 \\

40000
& 32.42
& 23.61
& 52.97
& 0.5538 \\

\rowcolor{cyan!10}
5000*
& 31.85
& 23.72
& 53.60
& \textbf{0.5564} \\

\rowcolor{cyan!10}
10000*
& 31.80
& \textbf{23.83}
& 54.10
& \textbf{0.5564} \\

\rowcolor{cyan!10}
20000*
& \textbf{32.91}
& 23.79
& \textbf{55.77}
& 0.5410 \\
\bottomrule
\end{tabular}
}
\caption{Ablation study of Preserver during Stage I Training. All configurations use the same Locator trained jointly on VOR and ROSE. Rows marked with $^\ast$ correspond to ROSE-only Preserver training, while unmarked rows use the VOR--ROSE mixture. $\mathrm{PSNR}_{\mathrm{bg}}^{\mathrm{ref}}$ measures fidelity to the reference video background over a fixed top-left crop occupying $\tfrac{1}{16}$ of the frame area.}
\label{tab:Preserver_comparison}
\end{table}

\section{Conclusions}
In this paper, we introduced DualEraser to tackle the fundamental semantic--pixel conflict that has  constrained joint video object and effect removal. At the condition level, we resolved modality dissonance by explicitly injecting effect semantics via Bipartite Text and the MCCE and by introducing the LD-CFG module to adaptively balance the relative dominance of text and mask guidance across diverse scenes. At the optimization level, our proposed architecture breaks the entanglement between semantic erasure and background preservation by assigning these competing objectives to specialized Locator and Preserver experts. Extensive experiments validate that DualEraser achieves state-of-the-art performance on standard benchmarks while delivering robust removal of complex physical effects in open-world videos.

{
\small
\bibliographystyle{ieeenat_fullname}
\bibliography{main}
}

\clearpage

\appendix
\section{Implementation and Evaluation Details}
\subsection{Training Data, Sampling, and Preprocessing}
Our training data are constructed from VOR-Train~\cite{fu2026effecterase} and ROSE-Train~\cite{miao2025rose}. The two datasets contain approximately 60K and 17K video pairs, respectively. ROSE-Train is highly imbalanced, with nearly 10K samples in the ``common'' category. We therefore rebalance it by oversampling each of the other four effect categories to approximately 8K training instances. This procedure preserves the original video set and only adjusts the sampling frequency of each category. We then combine the rebalanced ROSE training pool with VOR-Train, obtaining approximately 100K sampled video pairs in total. We refer to this construction as the \emph{balanced mixture of VOR and ROSE} in the main paper.

During Stage I, we adopt expert-specific training data to encourage the Locator and Preserver to learn their respective capabilities. For the 5B model, the Locator is trained for 12,500 steps on the balanced VOR--ROSE mixture, whereas the 1.3B Locator is trained for 15,000 steps on the same mixture. For both model sizes, the Preserver is trained for 20,000 steps exclusively on the rebalanced ROSE dataset. Stage I therefore comprises $32.5\mathrm{K}$ and $35\mathrm{K}$ expert-training steps for the 5B and 1.3B models, respectively. The mixed data expose the Locator to diverse objects, physical effects, and scene distributions, improving its ability to perform semantic erasure on in-the-wild videos. In contrast, the ROSE data provide the Preserver with supervision better suited to pixel-level background reconstruction.
As shown in \cref{tab:crop_ratio}, the real-world subset of VOR exhibits  larger background misalignment than ROSE.  We therefore train the Preserver on rebalanced ROSE to provide cleaner pixel-level supervision.

During Stage II, the Locator and Preserver are each further optimized for 1,800 steps on the balanced VOR--ROSE mixture. This corresponds to $3.6\mathrm{K}$ expert-training steps for both model sizes. Both experts use the same mixed-data distribution in this stage. This setting enables LD-CFG to learn conditional fusion across diverse synthetic and real-world scenes.

For data preprocessing, we apply affine transformations to the training videos to increase camera-viewpoint diversity. We also perform erosion and dilation on the input masks to improve the model's robustness to mask inaccuracies.

\begin{table}[t]
\centering
\scriptsize
\setlength{\tabcolsep}{4pt}
\renewcommand{\arraystretch}{1.5}
\resizebox{1.0\columnwidth}{!}{
\begin{tabular}{l|cc}
\toprule
\multirow[c]{2}{*}{\makecell{Training \\ Dataset}}
& Crop Ratio ($\tfrac{1}{16}$) & Crop Ratio ($\tfrac{1}{32}$) \\
& MAE$\downarrow$ & MAE$\downarrow$ \\
\hline\hline
ROSE (Synthetic) & 3.85 & 3.31 \\
VOR (Real) & 11.50 & 10.99 \\
\bottomrule
\end{tabular}
}
\caption{Background alignment of the VOR and ROSE training datasets. Five hundred reference--target pairs are sampled from each dataset, with VOR restricted to real-world data. Alignment is measured by the MAE between corresponding top-left crops at ratios of $\tfrac{1}{16}$ and $\tfrac{1}{32}$.}
\label{tab:crop_ratio}
\end{table}

\subsection{Bipartite Text Prompt Generation}
We use Qwen3-VL 8B~\cite{bai2025qwen3} to automatically generate the Bipartite Text prompts used for both training and evaluation. For each sample, Qwen3-VL receives two video inputs: (1) the reference video, which provides the complete scene context and contains the physical effects associated with the target object, and (2) a target-object-only video obtained by removing the background and retaining only the masked target region. The second input explicitly identifies the object to be removed, while the reference video enables the VLM to infer its associated effects and the surrounding scene context.

Based on these inputs, Qwen3-VL generates a Bipartite Text prompt containing two complementary components: removal semantics that specify the target object and its associated physical effects, and reconstruction semantics that describe the desired scene after removal. The instruction  used for this generation process is provided in Prompt~\ref{prompt:causal_effect}.

\prompt{Causal Effect Description}{
You are a video description expert responsible for describing the scene changes after a target object is removed. You will be provided with two videos: the first shows the scene before the target instances are removed, and the second is a foreground-only video that isolates the target instances to be removed.

First, you need to identify the causal effects of the instances, such as shadows, reflections, lighting, ripples, and other physical effects, in the first video. Then, you should write a prompt for the scene after the instances are removed. The description should follow the format below:

``After removing instances, their causal effects, such as shadows, reflections, emitted light, etc., should also be eliminated. The scene should appear as ...''

The description should describe the resulting video after removal, ensuring that it is consistent with physical laws and environmental realism.

Example 1:

After removing the desk lamp, its shadows and emitted light will disappear accordingly. Once the shadows are gone, the area previously covered by the shadow will reveal the desk's true white color; the area where the lamp was located will become darker, resulting in reduced overall illumination on the desk.

Example 2:

After removing the swan, its reflection on the water surface and the ripples it caused will also vanish. With the reflection gone, the area of the water surface that previously reflected the swan will return to the original greenish-blue water surface and its natural state. The overall scene will appear tranquil and harmonious, with the water surface showing continuous and gentle ripples.
}{prompt:causal_effect}

\subsection{Human Evaluation Protocol}

\paragraph{Evaluation setup.}
VOR-Wild~\cite{fu2026effecterase} contains $N=195$ input videos, and each evaluated method produces one result for every input. For each evaluation case, the outputs of all methods are anonymized and displayed in an independently randomized order. Raters may select multiple outputs in which the target object and its associated physical effects are successfully removed while satisfactory visual quality is preserved. A selected output receives a binary score of $1$, whereas an unselected output receives a score of $0$.

\paragraph{Erasure Preference Rate.}
For EPR, two experts in video object removal independently evaluate all 195 cases. Let $s_{m,i,r}\in\{0,1\}$ denote the score assigned to method $m$ for input $i$ by expert $r$. The EPR of method $m$ is computed as
\begin{equation}
\operatorname{EPR}_m
=
\frac{1}{2N}
\sum_{i=1}^{N}
\sum_{r=1}^{2}
s_{m,i,r}.
\end{equation}

\paragraph{User Preference Rate.}
For UPR, five volunteers follow the same evaluation protocol. Each volunteer evaluates a randomly assigned subset of 117 input cases, where each case includes the anonymized outputs of all evaluated methods. The assignments are balanced such that every input case is evaluated by exactly three distinct volunteers. Accordingly, the UPR of method $m$ is computed as
\begin{equation}
\operatorname{UPR}_m
=
\frac{1}{3N}
\sum_{i=1}^{N}
\sum_{r\in\mathcal{R}_i}
s_{m,i,r},
\end{equation}
where $\mathcal{R}_i$ denotes the three volunteers assigned to input $i$.

\section{Cross-Method Transferability of Bipartite Text}

In the main comparison of~\cref{main-tab:method_comparison}, we use Qwen3-VL 8B~\cite{bai2025qwen3} to generate method-specific prompts that follow the original prompt format of each baseline. To evaluate the direct transferability of our Bipartite Text prompt, we keep the pretrained checkpoints and other inference settings unchanged and replace only the original text condition with our prompt.

As shown in \cref{tab:bipartite_prompt_baseline}, direct prompt transfer affects the baselines to different degrees. EffectErase exhibits the largest degradation, with PSNR decreases of 6.06 dB on ROSE-Bench and 4.82 dB on VOR-Eval. ROSE also drops by 4.67 dB and 1.74 dB on the two benchmarks, respectively. SVOR experiences a  1.28 dB decrease on ROSE-Bench, although its PSNR on VOR-Eval changes only marginally. In contrast, Gen. Omnimatte remains nearly unchanged.

We attribute these differences to the distinct textual conditioning strategies used during training. For example, EffectErase~\cite{fu2026effecterase} replaces object-related text tokens with visual embeddings. This design reduces its compatibility with explicit descriptions of the target object, associated effects, and post-removal background. Gen. Omnimatte~\cite{lee2025generative}, by contrast, is more compatible with our prompt because its native conditioning already emphasizes background descriptions.

These results indicate that directly substituting the inference prompt is insufficient for models trained with different textual formats. Incorporating Bipartite Text during training or fine-tuning would be necessary to fairly assess whether these baselines can benefit from its additional effect and reconstruction semantics.

\begin{table}[t]
\centering
\tiny
\setlength{\tabcolsep}{1pt}
\renewcommand{\arraystretch}{1.5}
\resizebox{\linewidth}{!}{
\begin{tabular}{l|ccc|ccc}
\toprule
\multirow[c]{2}{*}{Method}
& \multicolumn{3}{c|}{ROSE-Bench}
& \multicolumn{3}{c}{VOR-Eval}\\
& PSNR$\uparrow$ & SSIM$\uparrow$ & LPIPS$\downarrow$
& PSNR$\uparrow$ & SSIM$\uparrow$ & LPIPS$\downarrow$ \\
\hline\hline
Gen. Omnimatte
& {27.07} & {0.8942} & 0.0856
& {21.62} & {0.7598} & {0.1557}\\
EffectErase
& 20.98 & 0.8679 & 0.1004
& 17.65 & 0.6993 & 0.1932\\
ROSE
& 26.42 & 0.8902 & {0.0770}
& 20.33 & 0.7378 & 0.1682\\
SVOR
& {29.89} & {0.9152} & {0.0571}
& {21.78} & {0.7552} & {0.1368}\\
\bottomrule
\end{tabular}
}
\caption{Inference-time transfer of Bipartite Text to baseline methods.}
\label{tab:bipartite_prompt_baseline}
\end{table}

\begin{figure}[t]
    \centering
    \includegraphics[width=1.0\linewidth]{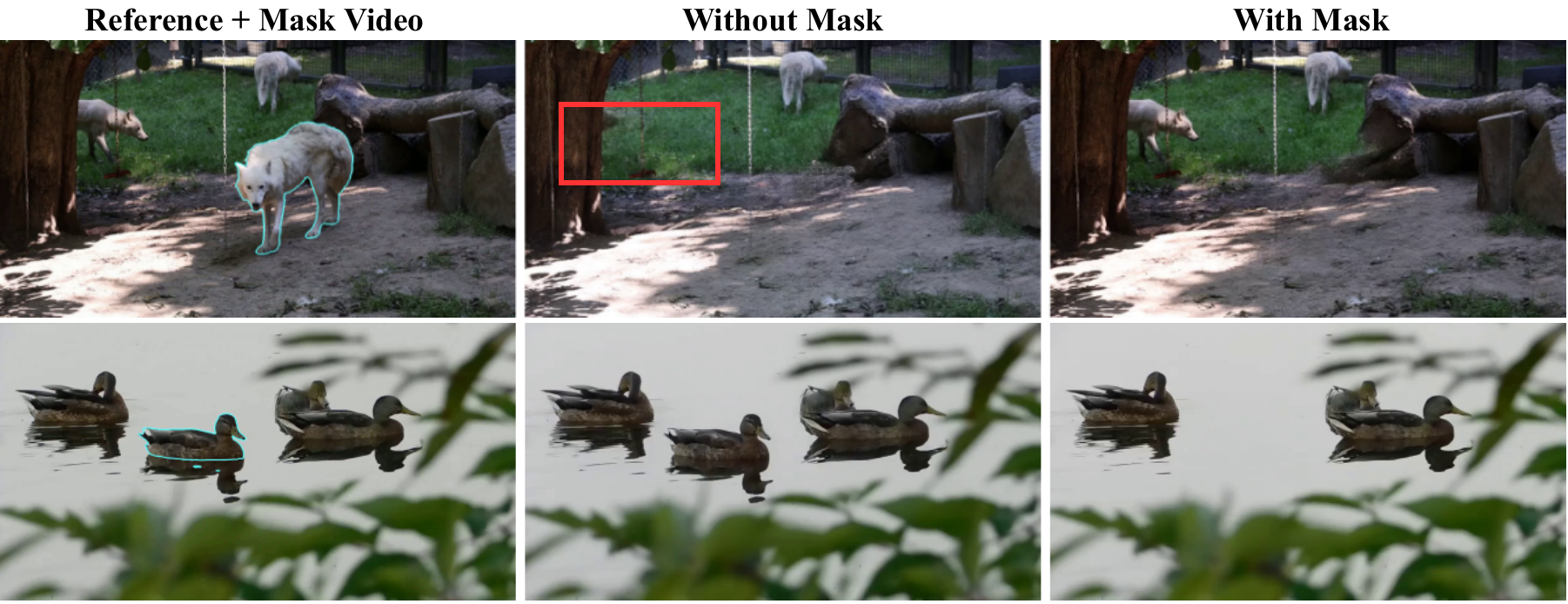}
    \caption{Qualitative comparison with and without spatial mask guidance. Without the mask condition, the model may fail to localize the intended target or modify unrelated regions.}
    \label{fig:no_mask}
    \vspace{0.7em}
    \tiny
    \setlength{\tabcolsep}{1pt}
    \renewcommand{\arraystretch}{1.2}
    \resizebox{\linewidth}{!}{
    \begin{tabular}{l|ccc|ccc}
    \toprule
    \multirow[c]{2}{*}{Mask Setting}
    & \multicolumn{3}{c|}{ROSE-Bench}
    & \multicolumn{3}{c}{VOR-Eval}\\
    & PSNR$\uparrow$ & SSIM$\uparrow$ & LPIPS$\downarrow$
    & PSNR$\uparrow$ & SSIM$\uparrow$ & LPIPS$\downarrow$ \\
    \hline\hline
    Without Mask
    & 30.32 & 0.9039 & 0.0687
    & 21.30 & 0.7376 & 0.1761\\
    \rowcolor{cyan!10}
    With Mask
    & 33.55 & 0.9262 & 0.0467
    & 23.87 & 0.7695 & 0.1197\\
    \bottomrule
    \end{tabular}
    }
    \captionsetup{hypcap=false}
    \captionof{table}{Comparison with and without spatial mask guidance.}
    \captionsetup{hypcap=true}
    \label{tab:mask_ablation}
\end{figure}
\section{Robustness and Ablation Studies}
\subsection{VLM Selection and Robustness}

We use Qwen3-VL 8B~\cite{bai2025qwen3} as the default prompt generator because it provides a practical balance between semantic quality and model size. To evaluate the sensitivity of DualEraser to the prompt-generating VLM, we additionally test InternVL3.5 1B~\cite{wang2025internvl3} and Qwen2.5-VL 3B~\cite{Bai2025Qwen25VLTR}. For all variants, we use the same video inputs and instruction template, while keeping the DualEraser checkpoint and all other inference settings unchanged.

As shown in \cref{tab:vlm_comparison}, changing the prompt-generating VLM produces only small variations in PSNR, SSIM, and LPIPS on ROSE-Bench and VOR-Eval. Qwen3-VL 8B achieves the highest EPR of 0.5846 on VOR-Wild, outperforming InternVL3.5 1B and Qwen2.5-VL 3B. These results indicate that DualEraser is compatible with different VLMs, while a stronger prompt generator can provide more accurate effect semantics and improve removal performance in open-world videos.

\begin{table*}[t]
\centering
\tiny
\setlength{\tabcolsep}{3pt}
\renewcommand{\arraystretch}{0.95}
\resizebox{0.9\linewidth}{!}{
\begin{tabular}{l|ccc|ccc|c}
\toprule
\multirow[c]{2}{*}{VLM}
& \multicolumn{3}{c|}{ROSE-Bench}
& \multicolumn{3}{c|}{VOR-Eval}
& VOR-Wild \\
& PSNR$\uparrow$ & SSIM$\uparrow$ & LPIPS$\downarrow$
& PSNR$\uparrow$ & SSIM$\uparrow$ & LPIPS$\downarrow$
& EPR$\uparrow$ \\
\hline\hline
InternVL3.5 1B
& 33.42 & 0.9255 & \underline{0.0467}
& \underline{23.91} & \underline{0.7701} & 0.1216
& \underline{0.5564}\\
Qwen2.5-VL 3B
& \underline{33.49} & \textbf{0.9269} & \textbf{0.0462}
& \textbf{23.92} & \textbf{0.7703} & \underline{0.1205}
& 0.5513\\
\hline
\rowcolor{cyan!10}
Qwen3-VL 8B
& \textbf{33.55} & \underline{0.9262} & \underline{0.0467}
& 23.87 & 0.7695 & \textbf{0.1197}
& \textbf{0.5846}\\
\bottomrule
\end{tabular}
}
\caption{Comparison of prompt-generating VLMs on ROSE-Bench, VOR-Eval, and VOR-Wild.}
\label{tab:vlm_comparison}
\end{table*}

\subsection{Robustness to Incomplete Bipartite Prompts}

VLM-generated prompts may omit individual semantic components. We therefore evaluate four incomplete text settings. ``Empty text'' removes all textual guidance. ``Object-only'' retains the target-object description, ``Effect-only'' retains the effect-removal instruction, and ``Background-only'' retains the desired post-removal scene description. All settings are derived from the complete Bipartite Text prompt while keeping the model checkpoint and other inference settings unchanged.

As shown in \cref{tab:text_setting_comparison}, the pixel-level metrics remain relatively stable across the different settings. However, the complete prompt achieves the highest VOR-Wild EPR of 0.5923, outperforming the incomplete settings. This indicates that the complete prompt primarily improves semantic removal in open-world videos.

The qualitative results in \cref{fig:no_perfect_text1}  illustrate the distinct roles of the different prompt components. When the effect-removal description is omitted, as in the ``Empty text'' and ``Object-only'' settings, the mask may still guide the removal of the target object, but weakly correlated effects such as diffuse smoke can remain. Conversely, when the post-removal background description is absent, the model receives insufficient guidance about the desired appearance of the restored region, which may result in inconsistent  scene structure. Notably, the ``Background-only'' setting can still eliminate smoke because its retained scene description explicitly specifies that no smoke should remain. The complete Bipartite Text combines explicit object-and-effect removal semantics with a description of the desired background, thereby producing more complete erasure and more coherent reconstruction.

\begin{table*}[!t]
\centering
\tiny
\setlength{\tabcolsep}{3pt}
\renewcommand{\arraystretch}{0.95}
\resizebox{0.90\linewidth}{!}{
\begin{tabular}{l|ccc|ccc|c}
\toprule
\multirow[c]{2}{*}{Prompt Setting}
& \multicolumn{3}{c|}{ROSE-Bench}
& \multicolumn{3}{c|}{VOR-Eval}
& VOR-Wild \\
& PSNR$\uparrow$ & SSIM$\uparrow$ & LPIPS$\downarrow$
& PSNR$\uparrow$ & SSIM$\uparrow$ & LPIPS$\downarrow$
& EPR$\uparrow$ \\
\hline\hline
Empty text
& 33.31 & 0.9250 & 0.0470
& 23.85 & 0.7687 & 0.1232
& 0.5231\\
Object-only
& 33.36 & 0.9252 & 0.0471
& 23.85 & \underline{0.7700} & \underline{0.1201}
& 0.5282\\
Effect-only
& 33.14 & \underline{0.9258} & 0.0470
& \underline{23.87} & 0.7697 & 0.1225
& 0.5256\\
Background-only
& \underline{33.47} & 0.9257 & \textbf{0.0466}
& \textbf{23.96} & \textbf{0.7709} & 0.1202
& \underline{0.5385}\\
\hline
\rowcolor{cyan!10}
Complete prompt
& \textbf{33.55} & \textbf{0.9262} & \underline{0.0467}
& \underline{23.87} & 0.7695 & \textbf{0.1197}
& \textbf{0.5923}\\
\bottomrule
\end{tabular}
}
\caption{Stress test of incomplete Bipartite Text settings on ROSE-Bench, VOR-Eval, and VOR-Wild. }
\label{tab:text_setting_comparison}

\vspace{1.5em}
\includegraphics[width=1.0\linewidth]{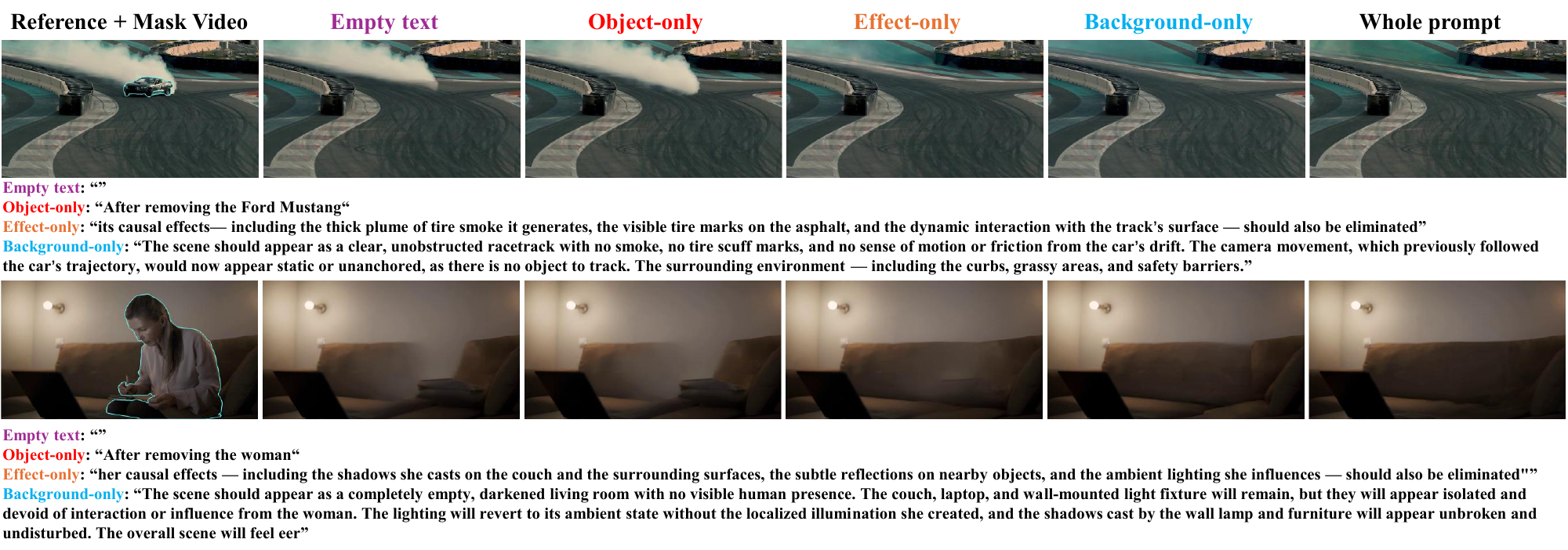}
\captionof{figure}{Qualitative stress test using incomplete Bipartite Text prompts. Missing effect-removal or background-reconstruction semantics can leave residual effects or produce inconsistent restored regions.}
\label{fig:no_perfect_text1}
\end{table*}

\begin{figure*}[!t]
\centering
\begin{subfigure}[t]{\columnwidth}
    \centering
    \includegraphics[width=0.90\linewidth]{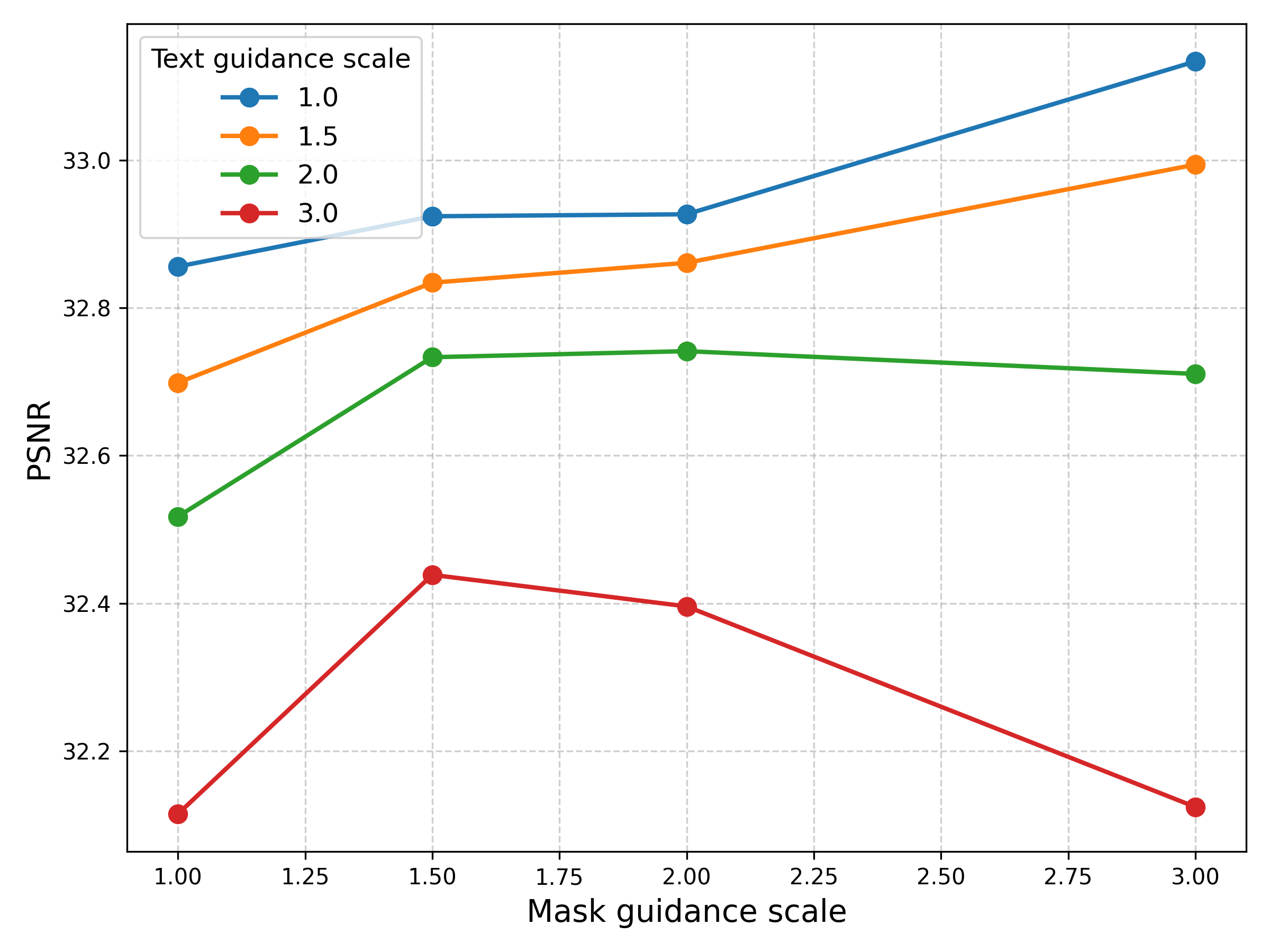}
    \caption{ROSE-Bench.}
    \label{fig:mc_cfg_rose}
\end{subfigure}\hfill
\begin{subfigure}[t]{\columnwidth}
    \centering
    \includegraphics[width=0.90\linewidth]{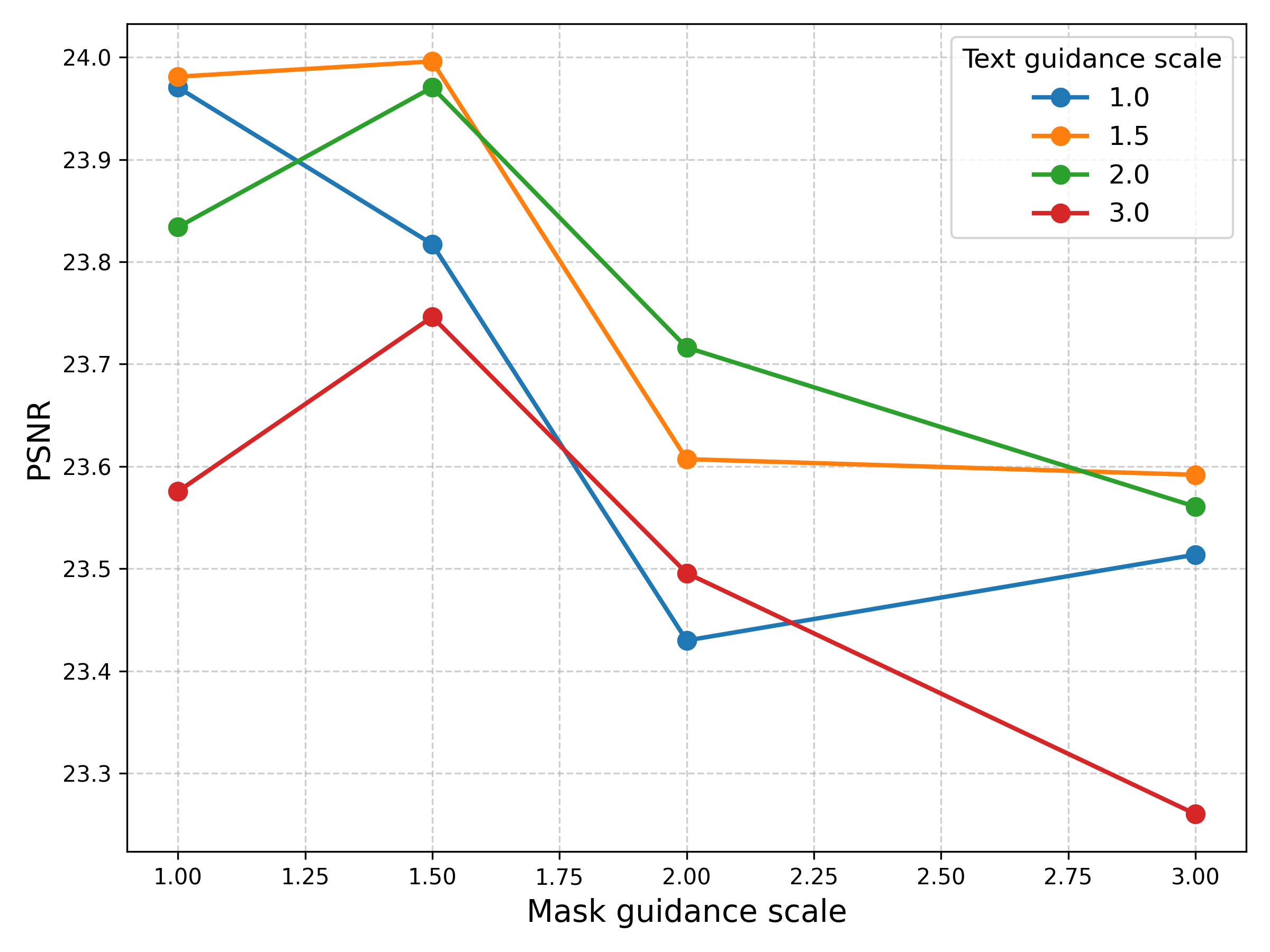}
    \caption{VOR-Eval.}
    \label{fig:mc_cfg_vor}
\end{subfigure}
\vspace{-0.5em}
\caption{Effect of the text and mask guidance scales on the PSNR of MC-CFG on (a) ROSE-Bench and (b) VOR-Eval.}
\label{fig:mc_cfg_guidance_scales}
\end{figure*}

\begin{table*}[!t]

\centering
\footnotesize
\setlength{\tabcolsep}{3.5pt}
\renewcommand{\arraystretch}{1.05}
\resizebox{0.92\linewidth}{!}{%
\begin{tabular}{c|ccc|ccc|ccc}
\toprule
\multirow{2}{*}{Steps}
& \multicolumn{3}{c|}{Boundary $=0.675$}
& \multicolumn{3}{c|}{Boundary $=0.775$}
& \multicolumn{3}{c}{Boundary $=0.875$} \\
& PSNR$\uparrow$ & SSIM$\uparrow$ & LPIPS$\downarrow$
& PSNR$\uparrow$ & SSIM$\uparrow$ & LPIPS$\downarrow$
& PSNR$\uparrow$ & SSIM$\uparrow$ & LPIPS$\downarrow$ \\
\hline\hline
1000* & 30.50 & 0.9127 & 0.0527 & 31.90 & 0.9220 & 0.0517 & 31.87 & 0.9199 & 0.0517 \\
2000* & 31.42 & 0.9198 & 0.0521 & \underline{32.18} & \textbf{0.9233} & 0.0523 & \underline{32.48} & \underline{0.9219} & \underline{0.0499} \\
3000* & \underline{32.22} & \underline{0.9209} & \underline{0.0493} & \textbf{32.49} & \underline{0.9232} & \underline{0.0515} & \textbf{32.51} & \textbf{0.9227} & 0.0501 \\
4000* & \textbf{32.74} & \textbf{0.9231} & \textbf{0.0492} & 31.29 & 0.9172 & \textbf{0.0512} & 32.22 & 0.9193 & \textbf{0.0479} \\
\bottomrule
\end{tabular}
}
\caption{Performance on  ROSE-Bench across Locator training durations and routing boundaries. * indicates ROSE-only training.}
\label{tab:boundary_ablation_rose}

\vspace{1.5em}
\centering
\footnotesize
\setlength{\tabcolsep}{3.5pt}
\renewcommand{\arraystretch}{1.05}
\resizebox{0.92\linewidth}{!}{%
\begin{tabular}{c|ccc|ccc|ccc}
\toprule
\multirow{2}{*}{Steps}
& \multicolumn{3}{c|}{Boundary $=0.675$}
& \multicolumn{3}{c|}{Boundary $=0.775$}
& \multicolumn{3}{c}{Boundary $=0.875$} \\
& PSNR$\uparrow$ & SSIM$\uparrow$ & LPIPS$\downarrow$
& PSNR$\uparrow$ & SSIM$\uparrow$ & LPIPS$\downarrow$
& PSNR$\uparrow$ & SSIM$\uparrow$ & LPIPS$\downarrow$ \\
\hline\hline
1000* & \underline{22.62} & 0.7602 & 0.1392 & 22.10 & 0.7624 & 0.1415 & 21.99 & 0.7625 & 0.1430 \\
2000* & 22.61 & \textbf{0.7675} & \textbf{0.1374} & 22.32 & \textbf{0.7659} & 0.1413 & \textbf{22.68} & \underline{0.7676} & \textbf{0.1362} \\
3000* & \textbf{22.63} & \underline{0.7673} & 0.1382 & \underline{22.42} & 0.7642 & \underline{0.1406} & \underline{22.59} & \textbf{0.7688} & 0.1375 \\
4000* & 22.44 & 0.7663 & \underline{0.1381} & \textbf{22.60} & \underline{0.7655} & \textbf{0.1386} & 22.23 & 0.7636 & \underline{0.1374} \\
\bottomrule
\end{tabular}
}
\caption{Performance on VOR-Eval across Locator training durations and routing boundaries.  * indicates ROSE-only training.}
\label{tab:boundary_vor}
\end{table*}

\begin{table}[htbp]
\centering
\tiny
\setlength{\tabcolsep}{3pt}
\renewcommand{\arraystretch}{0.95}
\resizebox{0.9\linewidth}{!}{
\begin{tabular}{l|ccc}
\toprule
\multirow[c]{2}{*}{\makecell{Locator \\ Training Steps}}
& \multicolumn{3}{c}{Routing Boundary} \\
& 0.675 & 0.775 & 0.875 \\
\hline\hline

1000* & \textbf{0.3897} & {0.3385} & \underline{0.3692} \\

2000* & \underline{0.3692} & \underline{0.3692} & \textbf{0.4128} \\

3000* & 0.2897 & \underline{0.3154} & \textbf{0.3769} \\

4000* & \underline{0.3385} & 0.3000 & \textbf{0.3795} \\

\bottomrule
\end{tabular}
}
\caption{VOR-Wild EPR under different Locator training durations and routing boundaries.  The asterisk (*) denotes ROSE-only training.}
\label{tab:boundary_vor_wild}
\vspace{-1.0em}
\end{table}

\subsection{Qualitative Robustness to Imperfect Masks and the Importance of Spatial Guidance}

The masks for all examples in
\cref{fig:show_smoke_image,fig:show_Deformation_image,fig:show_light_image,fig:show_Reflection_image} are automatically generated by
SAM3~\cite{carion2025sam}, whereas in \cref{fig:show_Mirror_image}, SAM3 is
used only for the first example. These automatically generated masks are not
always precise. In the first example of \cref{fig:show_Mirror_image}, the mask
contains  boundary inaccuracies. In the hand-removal example of
\cref{fig:show_Deformation_image}, the mask captures only part of the hand
and fails to cover the entire hand and arm. Despite these imperfections,
DualEraser correctly infers the intended target beyond the inaccurate mask
boundaries and removes the complete target together with its associated
effects.

We further evaluate the importance of spatial guidance by removing the mask
condition entirely while keeping the Bipartite Text prompt, model checkpoint,
and all other inference settings unchanged. The qualitative examples in
\cref{fig:no_mask} show that, without mask guidance, the model may fail to
localize the intended target, leave it partially or entirely unremoved, or
modify unrelated regions.

Quantitatively, as shown in \cref{tab:mask_ablation}, removing the mask
decreases PSNR from 33.55 to 30.32 dB on ROSE-Bench and from 23.87 to 21.30 dB
on VOR-Eval, corresponding to reductions of 3.23 and 2.57 dB, respectively.
Therefore, DualEraser tolerates moderate inaccuracies in automatically
generated masks, but spatial mask guidance remains important for precise and
localized erasure.

\subsection{Guidance-Scale Sensitivity of MC-CFG}

We evaluate MC-CFG using text and mask guidance scales selected from
$\{1.0,1.5,2.0,3.0\}$, while keeping the model checkpoint and all other
inference settings unchanged. As shown in
\cref{fig:mc_cfg_rose,fig:mc_cfg_vor}, the best-performing tested scale
combination in terms of PSNR differs across benchmarks. ROSE-Bench favors a
text guidance scale of $1.0$ and a mask guidance scale of $3.0$, whereas
VOR-Eval achieves its highest PSNR when both scales are set to $1.5$.

These results demonstrate that fixed MC-CFG is sensitive to manually selected
guidance scales and that a scale combination tuned for one data distribution
may not transfer optimally to another. This dataset-dependent sensitivity
motivates LD-CFG, which replaces manually specified global scales with
learnable feature fusion.

Together with the improvements over average and standard MC-CFG reported in
\cref{main-tab:deep_cfg}, these results support the use of learned conditional
fusion instead of manually tuned scalar guidance.

\subsection{Locator--Preserver Routing Boundary}
To facilitate faster convergence and reduce the computational cost of this
ablation, we train all variants exclusively on ROSE rather than on the
VOR--ROSE mixture. As reported in
\cref{tab:boundary_ablation_rose,tab:boundary_vor}, when the best score over
the evaluated Locator training durations is considered separately for each
metric, all three boundary values achieve comparable peak pixel-level
performance on both ROSE-Bench and VOR-Eval. As shown in
\cref{tab:boundary_vor_wild}, on VOR-Wild, a boundary
of 0.875 achieves the highest EPR at three of the four evaluated training
durations. We therefore adopt 0.875 as the default routing boundary.

\subsection{Failure Cases}

\begin{figure}[H]
    \centering
    \includegraphics[width=1.0\linewidth]{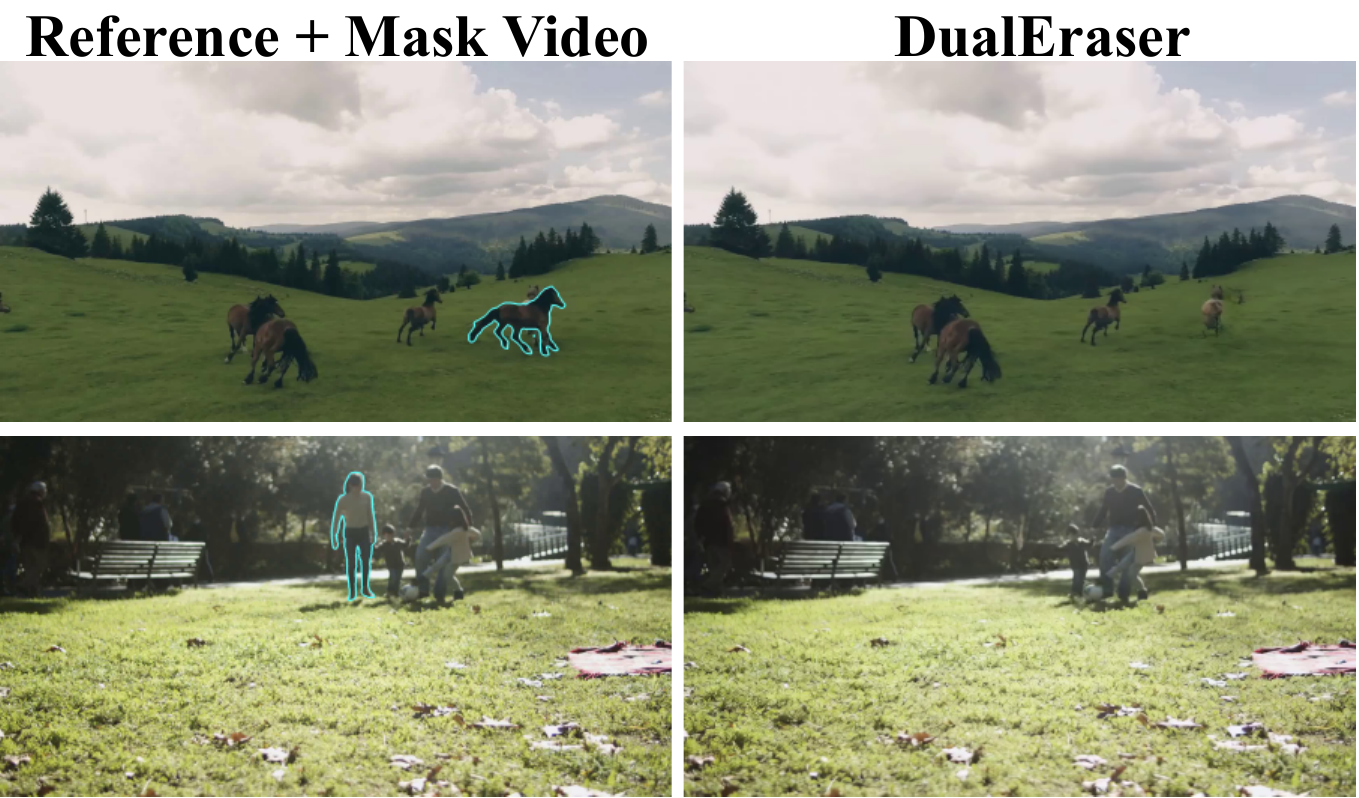}
    \caption{Failure cases involving occlusion reconstruction. After a removal target is erased, DualEraser may produce structural discontinuities when recovering another object that was partially occluded by the target.}
    \label{fig:bad_case}

    \includegraphics[width=1.0\linewidth]{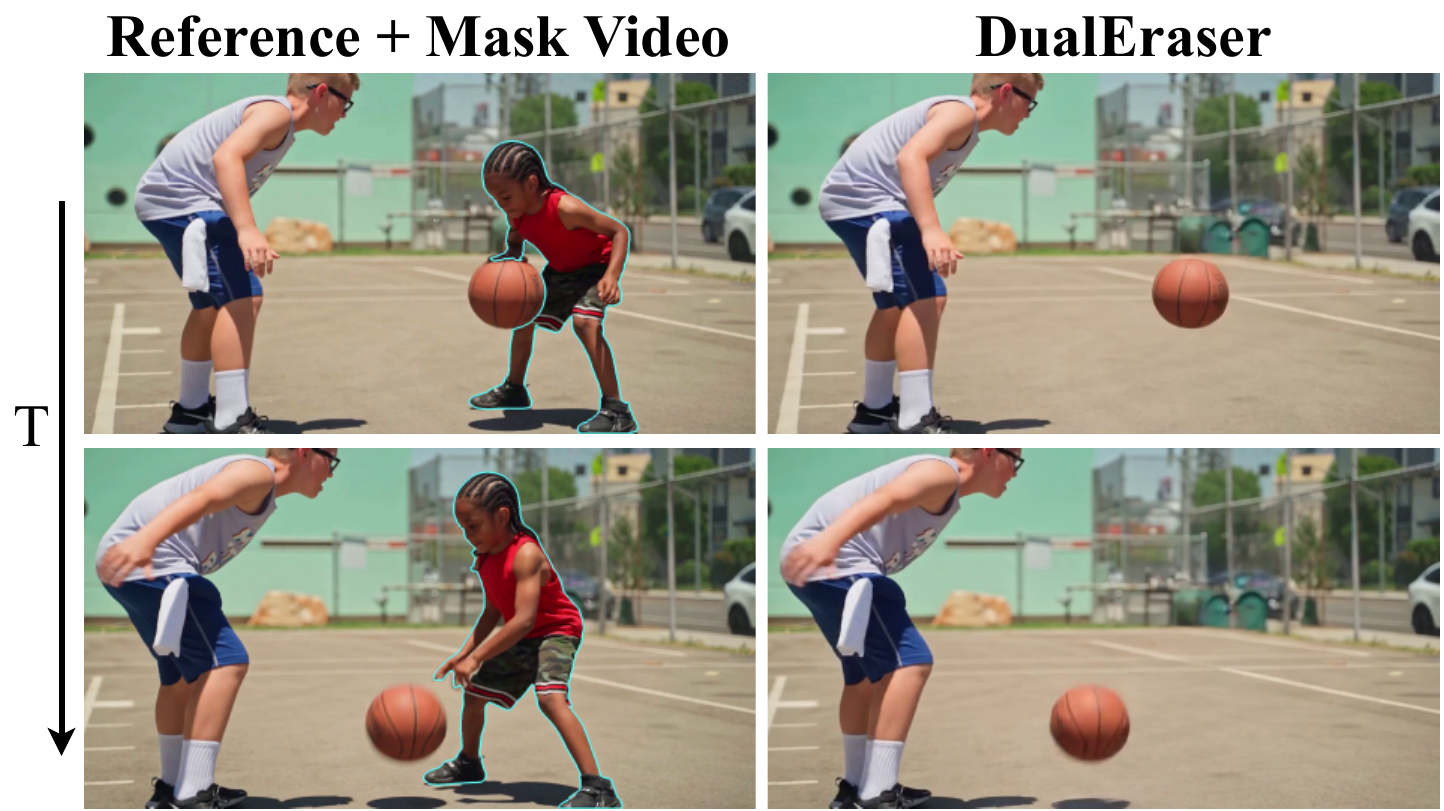}
    \caption{A failure case involving physically inconsistent causality. After the child is removed, the basketball retains nearly the same child-driven motion as in the reference video despite the absence of the corresponding physical interaction.}
    \label{fig:bad_case1}
\end{figure}

\paragraph{Occlusion Reconstruction.}
Reconstructing an object that is partially occluded by the removal target
remains challenging. As shown in \cref{fig:bad_case}, the target horse occludes
another horse behind it. After the target horse is removed, the model must
recover the unobserved body structure of the occluded horse from incomplete
visual evidence. In this case, DualEraser produces structural discontinuities
in the reconstructed horse.

\paragraph{Physically Inconsistent Causality.}
For the example shown in \cref{fig:bad_case1}, the basketball's motion is
caused by its interaction with the child. After the child is removed, the
basketball should no longer follow the same child-driven trajectory. However,
DualEraser preserves almost the same basketball motion as in the reference
video, even though the corresponding physical interaction and applied force
are no longer present. The resulting motion therefore lacks a valid physical
cause and violates physical laws.

These failures primarily stem from limitations in the generative capacity of the base model and are further exacerbated by the scarcity of similarly complex occlusion and physical-interaction scenarios in the training data.

\section{Additional Qualitative Comparisons}

\begin{figure*}[p]
    \centering
    \includegraphics[width=0.98\linewidth]{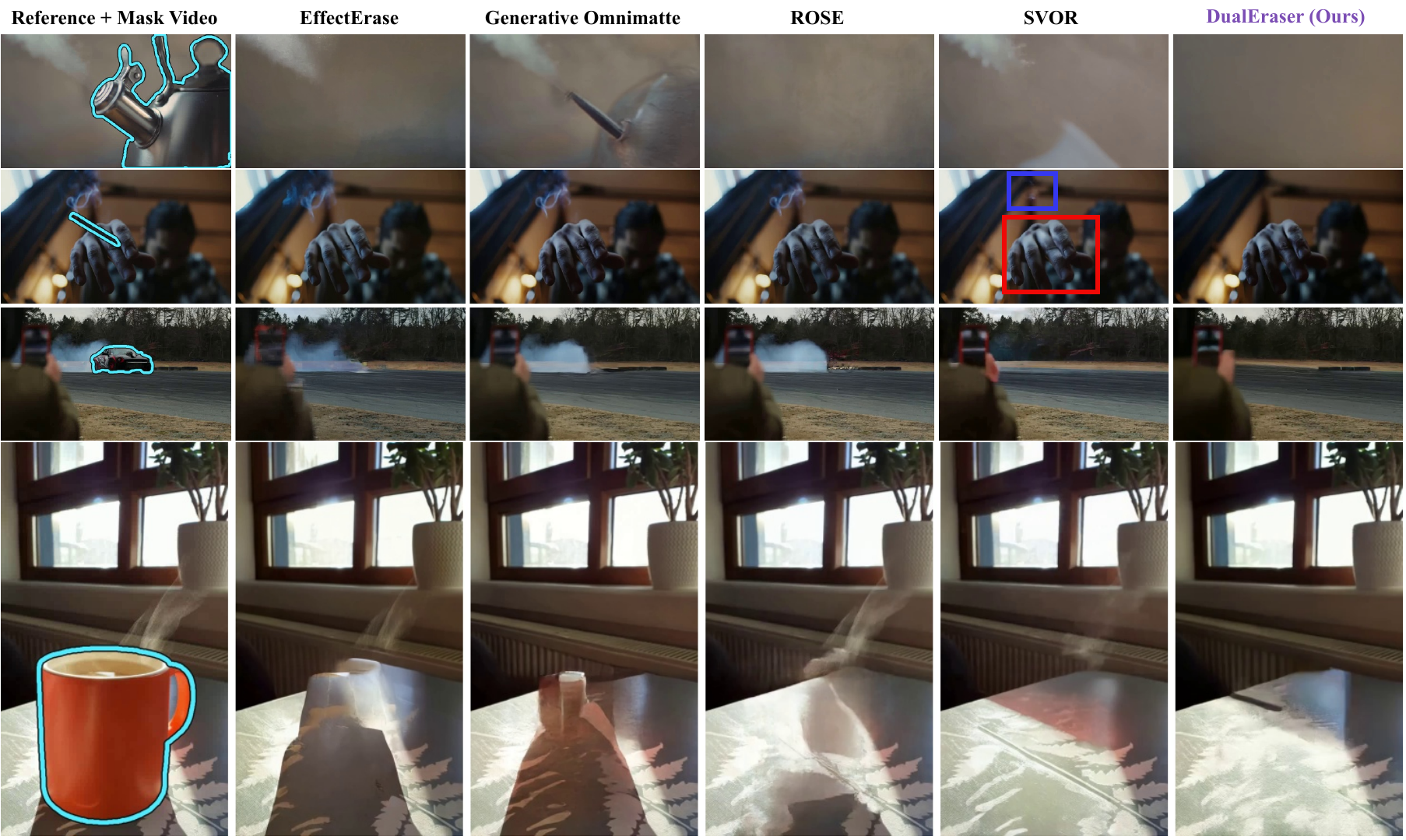}
    \caption{Qualitative comparisons for smoke removal. DualEraser removes target objects together with diffuse effects such as water vapor, cigarette smoke, and vehicle exhaust while preserving the surrounding scene.}
    \label{fig:show_smoke_image}

    \centering
    \includegraphics[width=0.98\linewidth]{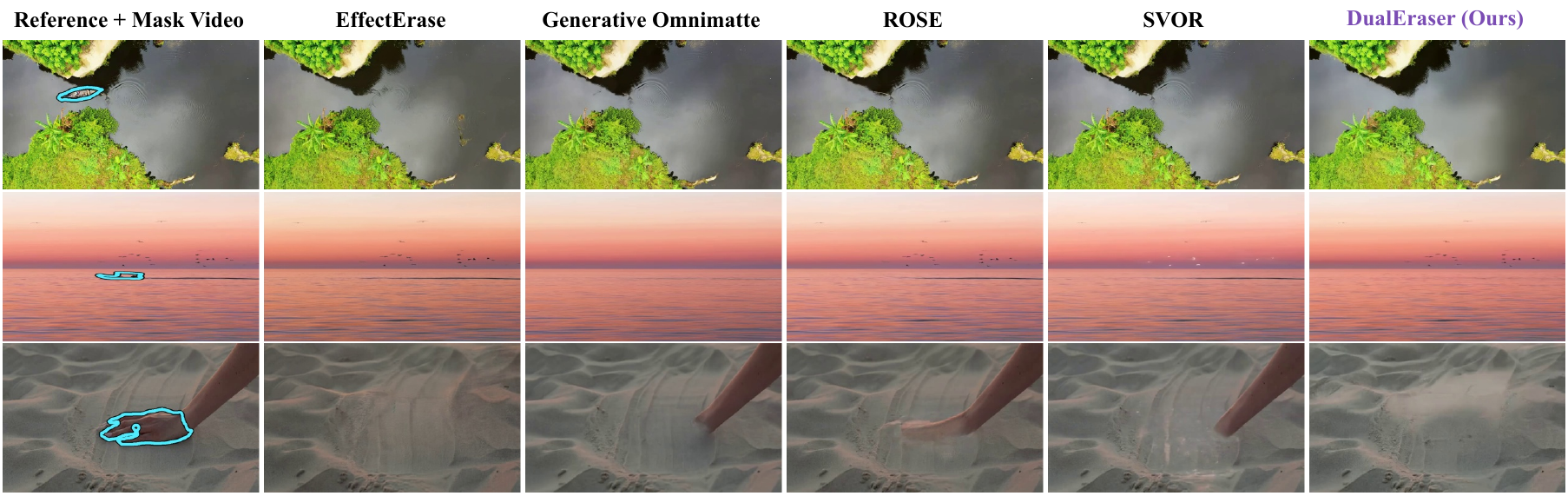}
    \caption{Qualitative comparisons for deformation removal. DualEraser removes target objects together with induced ripples and surface traces, producing more coherent water and sand backgrounds.}
    \label{fig:show_Deformation_image}

    \centering
    \includegraphics[width=0.98\linewidth]{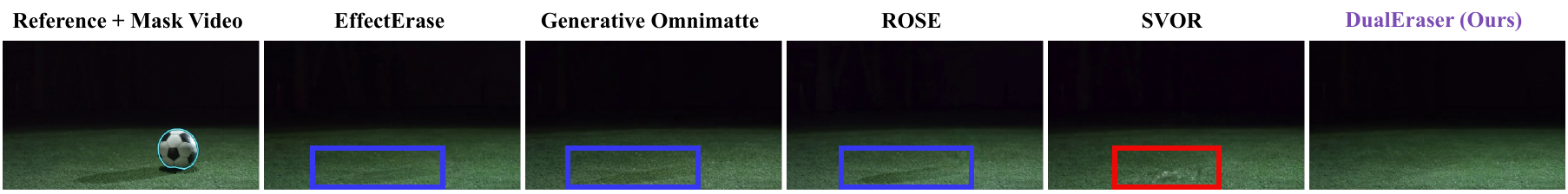}
    \caption{Qualitative comparisons for shadow removal. DualEraser removes target objects together with their cast shadows while reconstructing a coherent background.}
    \label{fig:show_shadow_image}
\end{figure*}

\begin{figure*}[p]
    \centering
    \includegraphics[width=0.98\linewidth]{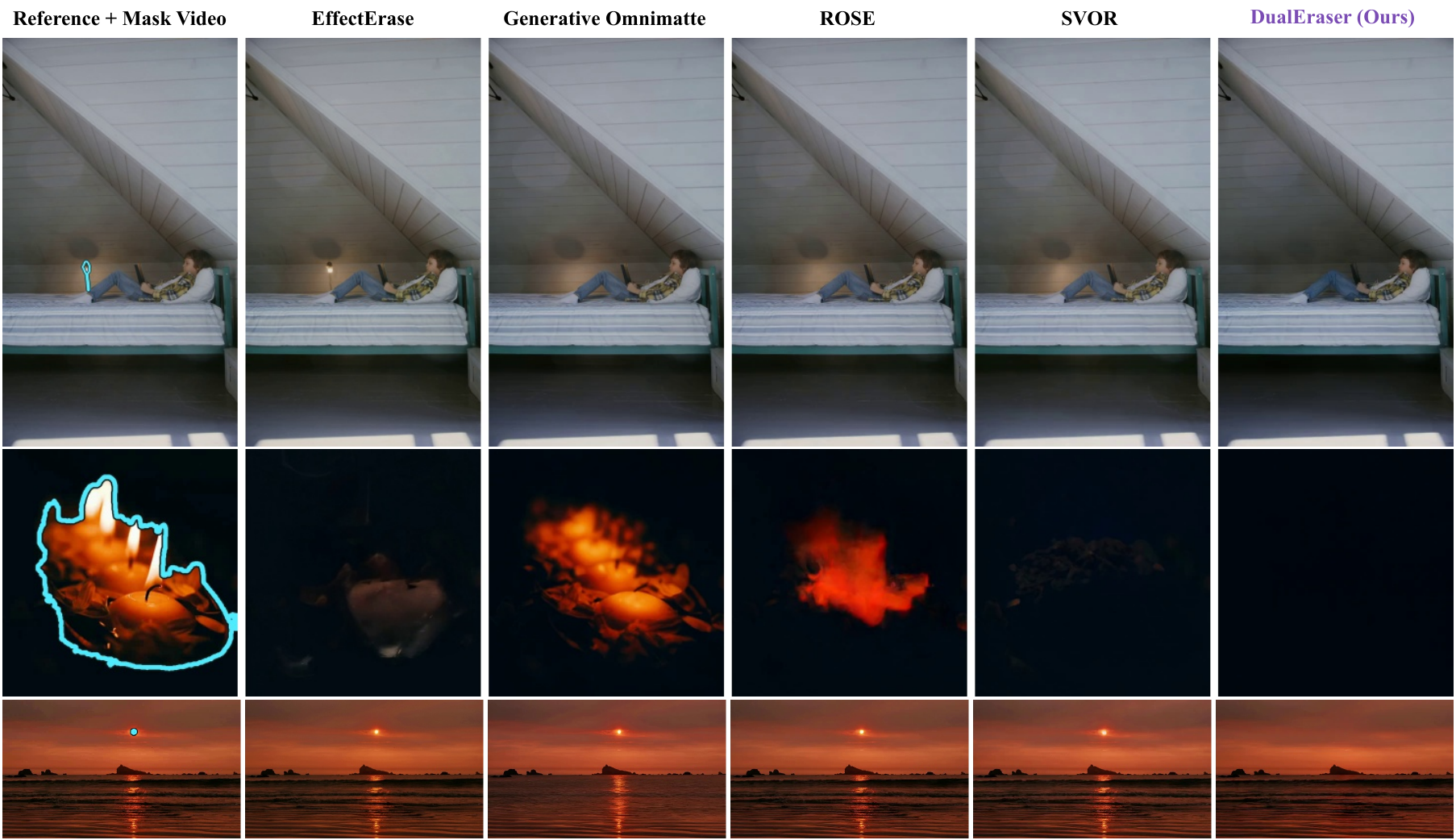}
    \caption{Qualitative comparisons for light removal. DualEraser removes light-emitting objects and their associated illumination while maintaining plausible scene lighting.}
    \label{fig:show_light_image}

    \centering
    \includegraphics[width=0.98\linewidth]{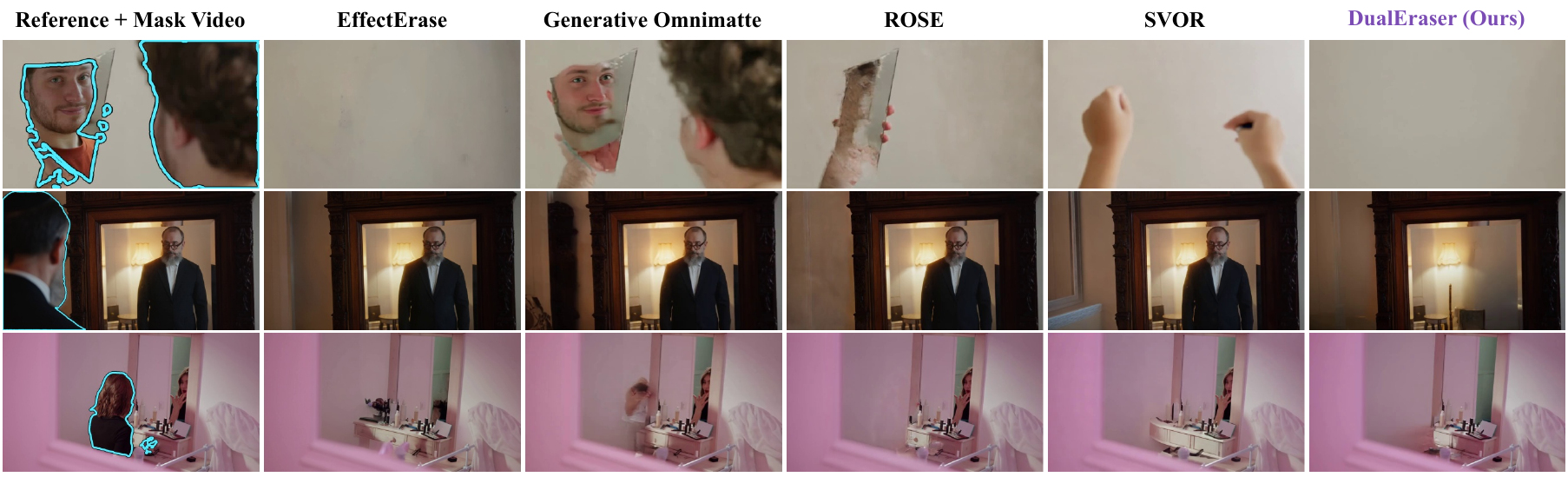}
    \caption{Qualitative comparisons for mirror removal. DualEraser removes both target objects and their mirror reflections while preserving the surrounding mirror structure.}
    \label{fig:show_Mirror_image}

    \centering
    \includegraphics[width=0.98\linewidth]{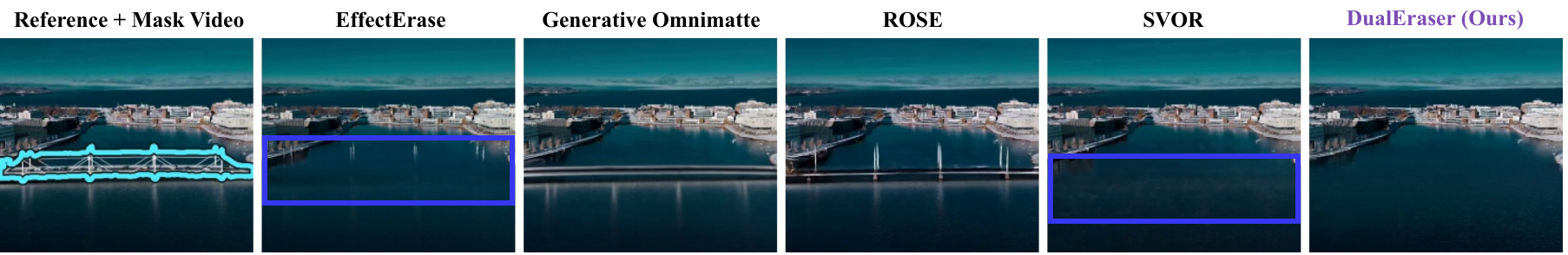}
    \caption{Qualitative comparisons for reflection removal. DualEraser removes target objects and their  reflections while reconstructing continuous water surfaces.}
    \label{fig:show_Reflection_image}
\end{figure*}

Additional qualitative comparisons are provided in \cref{fig:show_smoke_image,fig:show_Deformation_image,fig:show_light_image,fig:show_Mirror_image,fig:show_shadow_image,fig:show_Reflection_image}, covering diverse physical effects such as smoke, deformation,  light, mirror , shadows, and  reflections. Across these examples, competing methods may leave residual objects or effects and introduce background artifacts, whereas DualEraser more consistently removes both the target and its associated effects while reconstructing coherent backgrounds, including when the input masks are imperfect.

\clearpage
\makeatletter
\setlength{\@dblfptop}{0pt}
\makeatother

\clearpage

\end{document}